\documentclass[runningheads]{llncs}
\usepackage{graphicx}
\usepackage{csquotes}
\usepackage{tikz}
\usepackage{comment}
\usepackage{amsmath,amssymb} 
\usepackage{color}
\usepackage{etoolbox,lipsum}
\usepackage{color}
\usepackage{mathtools}
\usepackage{caption}
\usepackage{subcaption}
\usepackage{booktabs}
\usepackage{multirow}
\usepackage{epsfig}
\usepackage{multirow}
\usepackage{csquotes}
\usepackage{tikz}
\usepackage[pagebackref,breaklinks,colorlinks]{hyperref}
\usepackage[accsupp]{axessibility}  

\begin{document}
\pagestyle{headings}
\mainmatter
\def\ECCVSubNumber{6505}  


\title{Injecting 3D Perception of Controllable NeRF-GAN into StyleGAN for Editable \\Portrait Image Synthesis}


\titlerunning{Injecting 3D Perception of 3D-aware SURF-GAN into 2D StyleGAN} 
\authorrunning{J. Kwak et al.} 

\author{Jeong-gi Kwak$^{\dagger}$  \quad
Yuanming Li$^{\dagger}$ \quad
Dongsik Yoon$^{\dagger}$ \quad 
Donghyeon Kim$^{\dagger}$ \quad
David Han$^{\ddagger}$ \quad
Hanseok Ko$^{\dagger}$}
\institute{$^{\dagger}$ Korea University \qquad\qquad $^{\ddagger}$Drexel Univiersity
}

\maketitle

\begin{abstract}
Over the years, 2D GANs have achieved great successes in photorealistic portrait generation. However, they lack 3D understanding in the generation process, thus they suffer from multi-view inconsistency problem. To alleviate the issue, many 3D-aware GANs have been proposed and shown notable results, but 3D GANs struggle with editing semantic attributes. The controllability and interpretability of 3D GANs have not been much explored. In this work, we propose two solutions to overcome these weaknesses of 2D GANs and 3D-aware GANs. 
We first introduce a novel  3D-aware GAN, SURF-GAN, which is capable of discovering semantic attributes during training and controlling them in an unsupervised manner.
After that, we inject the prior of SURF-GAN into StyleGAN to obtain a high-fidelity 3D-controllable generator. 
Unlike existing latent-based methods allowing implicit pose control, the proposed 3D-controllable StyleGAN enables explicit pose control over portrait generation. 
This distillation allows direct compatibility between 3D control and  many StyleGAN-based techniques (e.g., inversion and stylization), and also brings an advantage in terms of computational resources.  
Our codes are available at \url{https://github.com/jgkwak95/SURF-GAN}.
\end{abstract}
\begin{figure}[!t]
\centering \includegraphics[width=\linewidth]{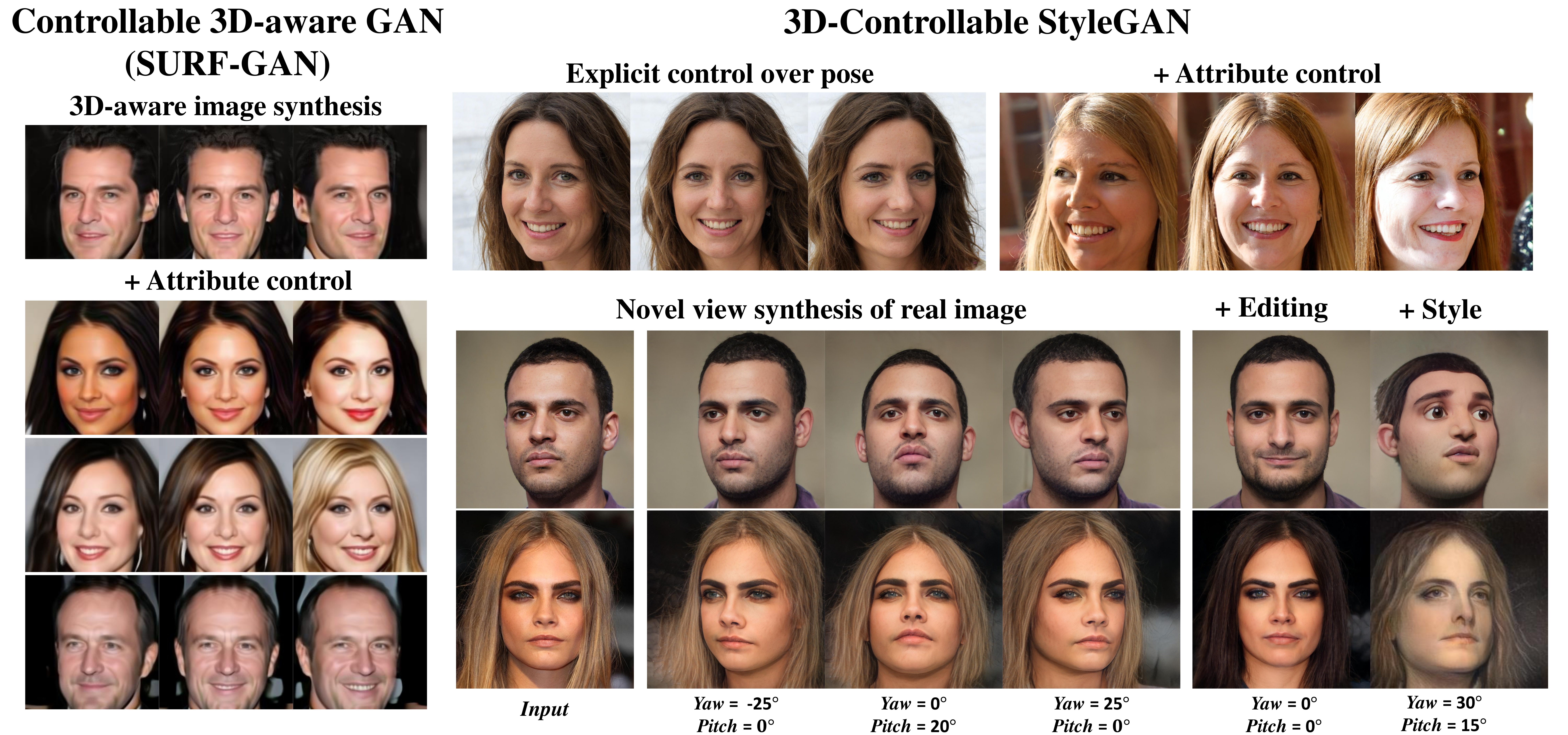}
\caption{\small{(Left): Proposed novel 3D-aware GAN (SURF-GAN) which is  capable of attribute-controllable generation as well as 3D-aware synthesis. (Right): 3D-controllable StyleGAN obtained by distilling the prior of SURF-GAN into 2D GAN.}}
\label{figure:thumbnail}

\end{figure} 


\section{Introduction}
Since the advent of Generative Adversarial Networks (GANs)~\cite{goodfellow2014generative}, 
remarkable progress has been made in the field of photorealistic image generation. 
The quality and diversity of images generated by 2D GANs have been improved considerably and recent models~\cite{karras2018progressive,brock2018large,karras2019style,karras2020analyzing,Karras2021} can produce high resolution images at a level that humans cannot distinguish. 
Despite the expressiveness of 2D GANs, they lack 3D understanding, in that the underlying 3D geometry of an object is ignored in the generation process. As a result, they suffer the problem of multi-view inconsistency. To overcome the issue, many researchers have studied 3D controllable image synthesis and it has become one of the mainstream research in the community.  
There have been several attempts to learn 3D pose information with 2D GAN by disentangling pose in the latent space, but they require auxiliary 3D supervision such as synthetic face dataset~\cite{KowalskiECCV2020} or 3DMM~\cite{deng2020disentangled,Tewari_2020_CVPR,tewari2020pie}. In addition, a few unsupervised approaches have been proposed by adopting implicit 3D feature~\cite{HoloGAN2019,BlockGAN2020} or differentiable renderer~\cite{shi2021lifting,pan2020gan2shape} in generation. However, these methods have struggled with multi-view consistency and photorealism.   

\begin{figure}[!t]
\centering \includegraphics[width=\linewidth]{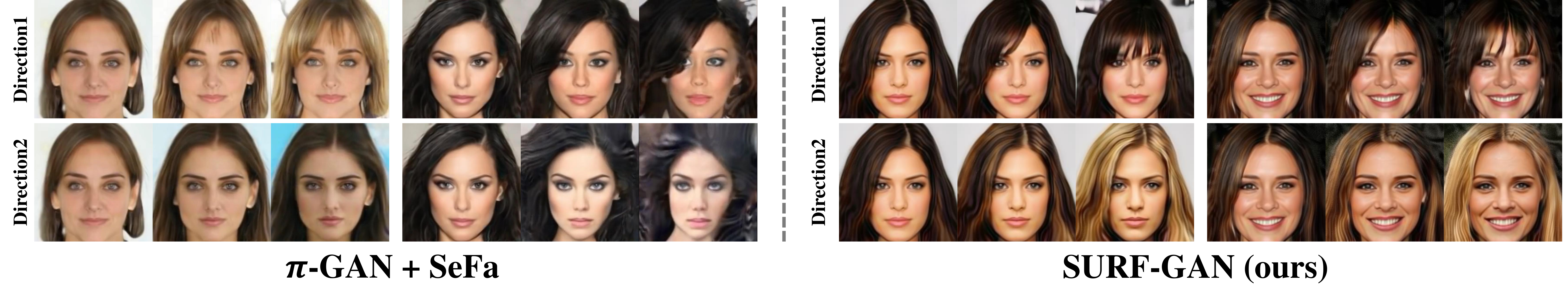}
\caption{\small{Results of attribute manipulation when applying SeFa~\cite{shen2021closedform} utilized for 2D GANs to a 3D-aware GAN~\cite{chan2021pi} (left). The captured attributes are entangled or not meaningful. In contrast, SURF-GAN can captures disentangled and semantic attributes (right).}}
\label{figure:pigan_semantic}

\end{figure}

Since the introduction of neural radiance fields (NeRF) by Mildenhall et al.~\cite{mildenhall2020nerf} which has achieved notable success in novel view synthesis, a new paradigm has emerged in 3D-aware generation, called 3D-aware GAN. 
Several researchers have proposed 3D-aware generative frameworks~\cite{Schwarz2020NEURIPS,Niemeyer2020GIRAFFE,chan2021pi,zhou2021CIPS3D,gu2021stylenerf,deng2022gram,xue2022giraffe} by leveraging NeRF as a 3D representation in GAN generator. 
NeRF-GANs learn 3D geometry from unlabelled images yet allow accurate and explicit control of 3D camera  based on a volume rendering. Despite the obvious advantages, 3D GANs based on a pure NeRF network require tremendous computational resources and  generate blurry images. Very recently, several approaches have alleviated the problems and have shown photorealistic output with high resolution by incorporating rear-end 2D networks~\cite{zhou2021CIPS3D,gu2021stylenerf,chan2022efficient,deng2022gram,xue2022giraffe}.
However 3D GANs have difficulty with attribute-controllable generation or real image editing because their latent space has been rarely investigated for interpretable generation (Fig.~\ref{figure:pigan_semantic}). 



In summary, these two distinct approaches have strengths and weaknesses that are complementary: 3D-aware GAN can generate novel poses but it has trouble with disentangling and manipulating attributes; 2D GAN is capable of controlling attributes but it struggles with 3D controllability.
In this work, we propose novel solutions to overcome each weakness of 2D GANs and 3D GANs.


First, we propose a novel 3D-aware GAN, i.e., SURF-GAN, which can discover semantic attributes by learning layer-wise \textbf{SU}bspace in INR Ne\textbf{RF}-based generator in an unsupervised manner.
The discovered semantic vectors can be controlled by corresponding parameters, thus this property allows us to  manipulate semantic attributes (e.g., gender, hair color, etc.) as well as explicit pose. 

With the proposed SURF-GAN, we take one more step to transform StyleGAN into a 3D-controllable generator.
We inject the prior of 3D-aware SURF-GAN into the expressive and disentangled latent space of 2D StyleGAN.
Unlike the previous methods~\cite{shen2020interpreting,harkonen2020ganspace,shen2021closedform} that allows implicit pose control, we make StyleGAN enable explicit control over pose. 
It means that the generator is capable of synthesizing accurate images based on a conditioned target view. 
By utilizing SURF-GAN which consists of pure NeRF layers as a generator of pseudo multi-view images,
the transformed StyleGAN can learn elaborate control over 3D camera pose with latent manipulation.
To this end, we proposed a method to find several orthogonal directions (not a single) related to the same pose attribute, and explicit control over the pose is accomplished by a combination of these directions. With a GAN inversion encoder, 3D controllable StyleGAN can be extended to the task of novel pose synthesis from a real image. 

In addition to 3D perception, we also inject the controllability about semantic attributes that SURF-GAN finds. 
We can find more pose-robust latent path in the latent space of StyleGAN because SURF-GAN can manipulate a specific semantic while keeping view direction unchanged. 
Moreover, it allows further applications related to StyleGAN family, e.g., 3D control over stylized images generated by fine-tuned StyleGAN.         
It is notable that our approach neither requires 3D supervision nor exploits auxiliary off-the-shelf 3D models (e.g., 3DMM or pose detector) in both training and inference because SURF-GAN learns 3D geometry from unlabelled 2D images from scratch. 


In summary, our contributions are as follow:
\begin{itemize}
    \item We propose a novel 3D-aware GAN, called SURF-GAN, which can discover controllable semantic attributes in an unsupervised manner.
    \item By injecting editing directions from the low-resolution 3D-aware GAN into the high-resolution 2D StyleGAN, we achieve a 3D controllable generator which is capable of explicit control over pose and 3D consistent editing.
    \item Our method is directly compatible with various well-studied 2D StyleGAN-based techniques such as inversion, editing or stylization. 
\end{itemize}

\section{Related Work}
\subsubsection{Pose-disentangled GANs.}
The remarkable advances have been achieved in photorealism by state-of-the-art GAN models~\cite{karras2018progressive,brock2018large,karras2019style,karras2020analyzing,Karras2021}. However, pose control by image generators has been limited due to a lack of 3D understanding in the synthesizing process.
Thereby, several works have attempted to disentangle the pose information from other attributes in 2D GANs.
The disentanglement has been achieved by leveraging supervision such as 3DMM~\cite{deng2020disentangled,Tewari_2020_CVPR,tewari2020pie,Yin_2017_ICCV}, landmark~\cite{hu2018pose}, synthetic images from 3D engine~\cite{KowalskiECCV2020} or pose detector~\cite{shoshan2021gan}.
A few unsupervised approaches without 3D supervision~\cite{HoloGAN2019,BlockGAN2020} have been proposed by disentangling pose with implicit 3D feature projection, but they allow only implicit 3D control and show blurry results.
Recently, a few methods~\cite{shi2021lifting,pan2020gan2shape} have incorporated a pre-trained StyleGAN with a differentiable renderer, but they struggle with photorealism, high-resolution~\cite{pan2020gan2shape} and real image editing~\cite{shi2021lifting}.  
%
\subsubsection{Interpretabilty and controllabiltiy of GAN.} The well-trained 2D GANs, such as StyleGAN~\cite{karras2019style,karras2020analyzing} have shown capable of disentangling the latent space. Recent works~\cite{shen2020interpreting,harkonen2020ganspace,abdal2021styleflow,nitzan2021large,shen2021closedform,yao2021latent,Patashnik_2021_ICCV} have demonstrated semantic manipulation, especially for facial attributes,
by analyzing the manifold and finding meaningful direction or mapping. Combining with GAN inversion~\cite{abdal2019image2stylegan,zhu2020domain,abdal2020image2stylegan++,richardson2021encoding,tov2021designing,roich2021pivotal,alaluf2021restyle,alaluf2021hyperstyle}, the applications of 2D GANs have been extended to real image editing. 
Alternatively, there have been studies~\cite{chen2016infogan,kaneko2017generative,lee2020high,jeon2021ib} that discover and disentangle latent embeddings into interpretable dimensions during training of the generator. EigenGAN~\cite{he2021eigengan} that inspired our approach has demonstrated interpretable latent dimensions by designing layer-wise subspace embedding. However, both types of methods support implicit control over the discovered semantics. 
In the case of a pose that can be defined with camera parameters, these methods struggle to synthesize explicit novel view elaborately.
Of course, the implicit methods can eventually create the desired pose through manual and iterative adjustment, but this is not an ideal situation. We can obtain a frontalized image automatically with some latent-based methods~\cite{shen2020interpreting,kwak2022generate,richardson2021encoding}, but not for arbitrary target pose. 
Recently, Chen et al.~\cite{chen2022sofgan} have introduced a generator allowing explicit control over pose, but it requires 3D mesh for pre-training process. 
\subsubsection{3D-aware GANs.}
Beyond the disentanglement of pose information, many efforts have been made to obtain 3D-awareness in generation. 
Earlier methods have adopted several explicit 3D representations in 2D image generation such as voxel~\cite{VON,lunz2020inverse,henzler2019platonicgan,gadelha20173d} or mesh~\cite{liao2020towards,szabo2019unsupervised}. However, they suffer from a lack of visual quality and limited resolution. Recently, approaches~\cite{Schwarz2020NEURIPS,chan2021pi,Niemeyer2020GIRAFFE,gu2021stylenerf,zhou2021CIPS3D,chan2022efficient,deng2022gram,or2022stylesdf,xue2022giraffe} based on neural fields have made significant progress in photorealism and 3D consistency. 
Nevertheless, these 3D-aware GANs have weakness in finding and editing semantic attribute because their latent space has been rarely investigated. 
Very recently, Sun et al.~\cite{sun2021fenerf} have proposed an editable NeRF-GAN, but it does not handle diverse semantic attributes and requires semantic maps as supervision. 
In addition, 3D GANs struggle with novel pose generation of real image despite their capability of multi-view consistency. Recently proposed EG3D~\cite{chan2022efficient} has shown experiments of novel view synthesis and presented outstanding results, but it requires iterative optimization for latent code and fine-tuning of the
generator~\cite{roich2021pivotal} for each target image.
           


\section{Proposed method} \label{sec:sec3}
In this section, we describe our method, by first introducing SURF-GAN in detail and then by explaining a method to inject the prior of 3D SURF-GAN into 2D StyleGAN. 
Note that the word \enquote{StyleGAN}  denotes StyleGAN2~\cite{karras2020analyzing}.

\subsection{Towards controllable NeRF-GAN} \label{sec:sec3.1}

\subsubsection{Preliminaries: NeRF-GANs.} 
Existing 2D GANs (e.g., StyleGAN~\cite{karras2019style,karras2020analyzing}) synthesize output image directly with sampled latent vector.
However, NeRF-GANs~\cite{chan2021pi,Niemeyer2020GIRAFFE,gu2021stylenerf,zhou2021CIPS3D} generate a radiance field~\cite{mildenhall2020nerf} before rendering 2D image. 
Given a position $\mathbf{x}\in\mathbb{R}^3$ and a viewing direction $\mathbf{v}\in\mathbb{S}^2$, 
it predicts a volume density $\mathbf{\sigma}(\mathbf{x})\in\mathbb{R}_{+}$ and the view-dependent RGB color $\mathbf{c}(\mathbf{x}, \mathbf{v})\in\mathbb{R}^3$ of the input point.
The points are sampled from rays of camera, and then an image is rendered into 2D grid with a classic volume rendering technique~\cite{kajiya1984ray}. 
To produce diverse images, existing NeRF-GAN methods adopt StyleGAN-like modulation, where some components in the implicit neural network, e.g., intermediate features~\cite{chan2021pi,zhou2021CIPS3D} or weight matrices~\cite{gu2021stylenerf}  are modulated by sampled noise passing though a mapping network. 
Thereby, NeRF-GAN can control the pose by manipulating viewing direction $\mathbf{v}$ and change identity by injecting different noise vector.
Nevertheless, it is ambiguous how to interpret the latent space and how to disentangle semantic attributes of NeRF-GAN for controllable image generation.

\begin{figure}[!t]
\centering \includegraphics[width=\linewidth]{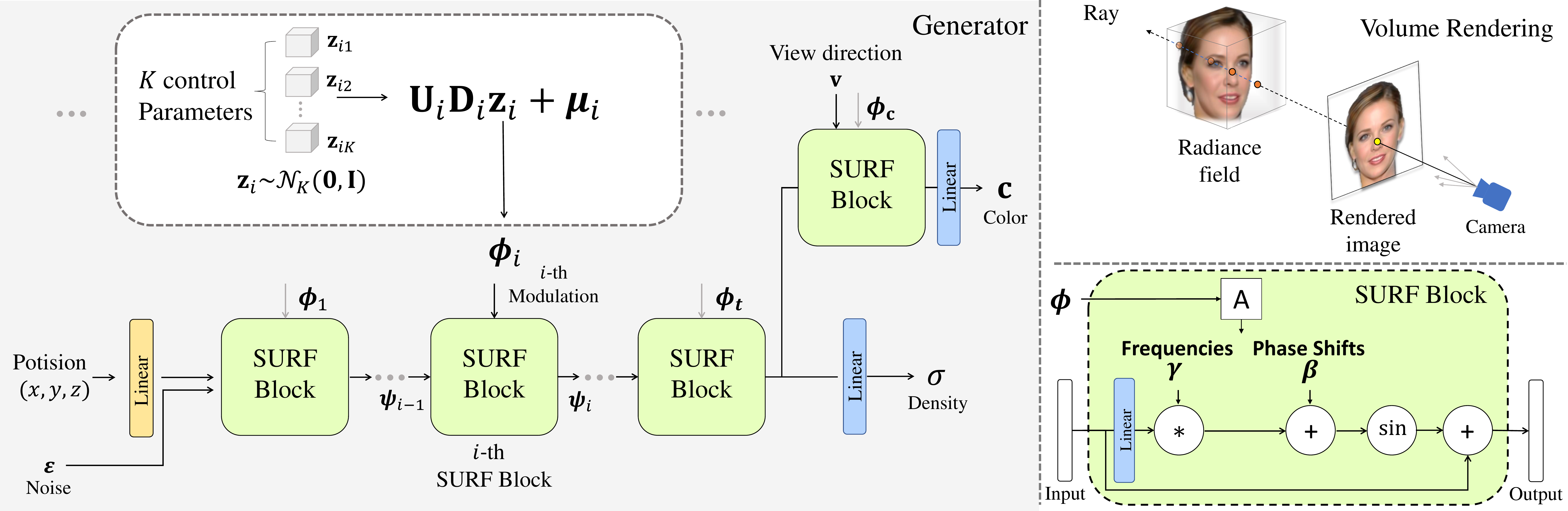}
\caption{\small{Overview of SURF-GAN generator. Interpretable dimensions are caputured in layers with sub-modulation vectors. As like INR 3D-aware GANs, it takes position and view direction as input and predicts view dependent color ($\mathbf{c}$) and its density ($\sigma$).}}   
\label{figure:SURFGAN} 
\end{figure} 
\subsubsection{Learning layer-wise subspace in NeRF network.}
Inspired by EigenGAN~\cite{he2021eigengan}, we adopt a different strategy from 
the existing methods~\cite{chan2021pi,zhou2021CIPS3D} those modulation is obtained by the mapping network consisting of several MLPs. EigenGAN learns interpretable subspaces in layers of its generator during training. However, EigenGAN is typical 2D convolution-based GAN framework, thus its concept is inapplicable to INR based NeRF-GAN. 
Therefore, we propose a novel framework (i.e., SURF-GAN), which captures the disentangled attributes in layers of NeRF network.  
Fig.~\ref{figure:SURFGAN} shows the overview of SURF-GAN. 
The generator consists of $t+1$ SURF blocks ($t$ for shared layers and one for color layer). Following $\pi$-GAN, SURF block adopts the feature-wise linear modulation (FiLM)~\cite{perez2018film} to transform the intermediate features with frequencies $\boldsymbol{\gamma}_{i}$ and phase shifts $\boldsymbol{\beta}_{i}$, and followed SIREN activation~\cite{sitzmann2020implicit}. SURF block in $i^{\text{th}}$ layer is formulated as
\begin{equation}
         \boldsymbol{\psi}_{i}=\text{SURF}_{i}\left( \boldsymbol{\psi}_{i-1}, \boldsymbol{\phi}_{i}\right)=\sin \left( \boldsymbol{\gamma}_{i} \cdot\left(\mathbf{W}_{i}  \boldsymbol{\psi}_{i-1}+\mathbf{b}_{i}\right)+ \boldsymbol{\beta}_{i}\right)+ \boldsymbol{\psi}_{i-1},
\end{equation}
where $ \boldsymbol{\psi}_{i-1}$ and $\boldsymbol{\phi}_{i}$ denote input feature and modulation of $i^{th}$ layer respectively. $\mathbf{W}_{i}$ and $\mathbf{b}_{i}$ represent the weight matrix and followed bias. Unlike other NeRF-GANs, we add skip connection~\cite{he2016deep} to prevent drastic change of modulation vectors in training.
In the model, a subspace embedded in each layer determines the modulation. Each subspace has orthogonal basis and it can be updated during training. The basis are learned to capture semantic modulation.
Concretely, in the case of $i^{th}$ layer, a specific subspace determines the modulation of $i^{th}$ layer of NeRF network. 
It consists of learnable matrices, orthonormal basis $\mathbf{U}_{i}=\left[\mathbf{u}_{i1}, \ldots, \mathbf{u}_{iK}\right]$ and a diagonal matrix $\mathbf{D}_{i}=\operatorname{diag}\left(d_{i1}, \ldots, d_{iK}\right)$. Each column of $\mathbf{U}_{i}$ plays a role of sub-modulation and it is updated to discover a meaningful direction that results in semantic change in image space. $d_{i1}, \ldots, d_{iK}$ serve as scaling factors of corresponding basis vectors $\mathbf{u}_{i1}, \ldots, \mathbf{u}_{iK}$. The latent $\mathbf{z}_i\in\mathbb{R}^K$ is set of $K$ scalar control parameters, i.e., 
\begin{equation} \label{eq:eq2}
\mathbf{z}_i = \left\{z_{ij} \in \mathbb{R} \mid z_{ij} \sim \mathcal{N}(0, 1), j=1, \ldots, K\right\},    
\end{equation}
where ${z}_{ij}$ is a coefficient of sub-modulation $d_{ij}\mathbf{u}_{ij}$. Hence, the modulation of $i^{th}$ layer $\phi_{i}$ is decided by weighted summation of $K$ sub-modulations with $\mathbf{z}_{i}$, i.e,
\begin{equation} \label{eq:eq3}
  \boldsymbol{\phi}_{i}=\mathbf{U}_{i} \mathbf{D}_{i} \mathbf{z}_{i}+ \boldsymbol{\mu}_{i}=\sum_{j=1}^{K} z_{ij}d_{ij}\mathbf{u}_{ij} + \boldsymbol{\mu}_{i},
\end{equation}
where the marginal vector $\boldsymbol{\mu}_{i}$ is employed to capture shifting bias. Finally, a simple affine transformation 
is applied to $\boldsymbol{\phi}_{i}$ for matching dimension and obtaining frequency $\boldsymbol{\gamma}_{i}$ and phase shift $\boldsymbol{\beta}_{i}$. At training phase, SURF-GAN layers learn variations of meaningful modulation controlled by randomly sampled $\mathbf{z}$. Additionally, an input noise $\epsilon$ is also injected to capture the rest variations missed by the layers.   
To improve the disentanglement of attributes and to prevent the basis fall into a trivial solution, we adopt the regularization loss to guarantee the column vectors of $\mathbf{U}_i$ to be orthogonal following EigenGAN, i.e., 
\begin{equation}
     \mathcal{L}_{\text{reg}} = \mathbb{E}_{i}[ \left\|\mathbf{U}_{i}^{\mathrm{T}} \mathbf{U}_{i}-\mathbf{I}\right\|_{1}].
\end{equation}
Finally, output image is rendered by volume rendering technique~\cite{kajiya1984ray}.   
At inference phase, we can control the discovered semantic attributes by manipulating corresponding element in $\mathbf{z}$. In addition, SURF-GAN enables explicit control over pose using viewing direction $\mathbf{v}$ as like other NeRF-based models.

\subsection{Explicit control over pose with StyleGAN} \label{sec:sec3.2}

In Sec.~\ref{sec:sec3.2} and Sec.~\ref{sec:sec3.3}, we introduce a method to inject 3D perception and attribute controllability of SURF-GAN into StyleGAN. 

\subsubsection{Leveraging 3D-aware SURF-GAN.}
The first step is to transform pre-trained StyleGAN into a 3D controllable generator. 
 We start with a question: How can we make StyleGAN be capable of controlling over pose explicitly when given arbitrary latent code? 
To this end, we utilize SURF-GAN as a pseudo ground-truth generator. It provides three images, i.e., $I_s$, $I_c$, $I_t$ which denote source, canonical, and target image respectively. 
Here, $\mathbf{z}$ is fixed in all images but the view directions of $I_s$ and $I_t$ are randomly sampled and $I_c$ has canonical view (i.e., $\mathbf{v}$=$[0, 0]$). Therefore, we can exploit them as multi-view supervision of the same identity. Afterwards, the images are embedded to $\mathcal{W}+$~\cite{abdal2019image2stylegan} space by a GAN inversion encoder $E$, i.e, 
$\{\mathbf{w}_s, \mathbf{w}_c, \mathbf{w}_t\} = \{E(I_s), E(I_c), E(I_t)\}$.
Here, we exploit the pre-trained pSp~\cite{richardson2021encoding} encoder and it actually predicts the residual and adds it to the mean latent vector, but we omit the notation for simplicity. 

\subsubsection{Mapping to a canonical latent vector.}
To handle arbitrary pose without employing off-the-shelf 3D models, we need to build an additional process.
To this end, we propose a canonical latent mapper $T$, which converts an arbitrary code to a canonical code in the latent space of StyleGAN. Here, the canonical code implies being a canonical pose (frontal) in image space. $T$ takes $\mathbf{w}_s$ as input and predicts its frontalized version $\mathbf{\hat{w}}_c=T(\mathbf{w}_s)$ with the mapping function. In order to train $T$, we exploit latent loss to minimize the difference between the predicted $\hat{\mathbf{w}}_c$ and pseudo ground truth of canonical code $\mathbf{w}_c$, i.e.,  
\begin{equation} \label{eq:eq}
    \mathcal{L}^{c}_{\text{w}} = \lVert{\mathbf{w}_c-T(\mathbf{w}_s)}\rVert_1.  
\end{equation}  
To guarantee plausible translation result in image space, we also adopt pixel-level $\ell_2$-loss and  LPIPS loss~\cite{zhang2018unreasonable} between two decoded images, i.e.,
\begin{align}
    \mathcal{L}^{c}_{\text{I}} &= \lVert{I'_{c}-\hat{I}_{c}}\rVert^2_2 \\
    \mathcal{L}^{c}_{\text{LPIPS}} &= \lVert{F(I'_{c})-F(\hat{I}_{c})}\rVert^2_2,
\end{align}
where $I'_{c}$ and $\hat{I}_{c}$ represent the decoded images from  
$\mathbf{w}_c$ and $\mathbf{\hat{w}}_c$ respectively, and $F(\cdot)$ denotes the perceptual feature extractor.
Hence, the loss for canonical view generation is formulated by
\begin{equation} \label{eq:eq9}
    \mathcal{L}^{\text{c}} = 
    \lambda_{1}\mathcal{L}^{\text{c}}_{\text{w}} + \lambda_{2}\mathcal{L}^{\text{c}}_{\text{I}} + \lambda_{3}\mathcal{L}^{\text{c}}_{\text{LPIPS}}. 
\end{equation}
\subsubsection{Target view generation.} Next, the canonical vector is converted to a target latent vector according to given a target view $\textbf{v}_t=\left[\alpha, \beta\right]$ as an additional input. Here, $\alpha$ and $\beta$ stand for pitch and yaw respectively. The manipulation is conducted in the latent space of StyleGAN by adding a pose vector which is obtained by a linear combination of pitch and yaw vectors ($\mathbf{p}$ and $\mathbf{y}$, respectively) with $\textbf{v}_t$ as coefficients, i.e., $\hat{\mathbf{w}}_{t}=\hat{\mathbf{w}}_c+\mathbf{L}\mathbf{v}_{t}^T$, where $\mathbf{L}=\left[\mathbf{p}\;\mathbf{y}\right]$. Therefore, we need to find optimal solution of $\mathbf{L}$ which can represent an adequate 3D control over pose. 
\begin{figure}[!t]
\centering \includegraphics[width=\linewidth]{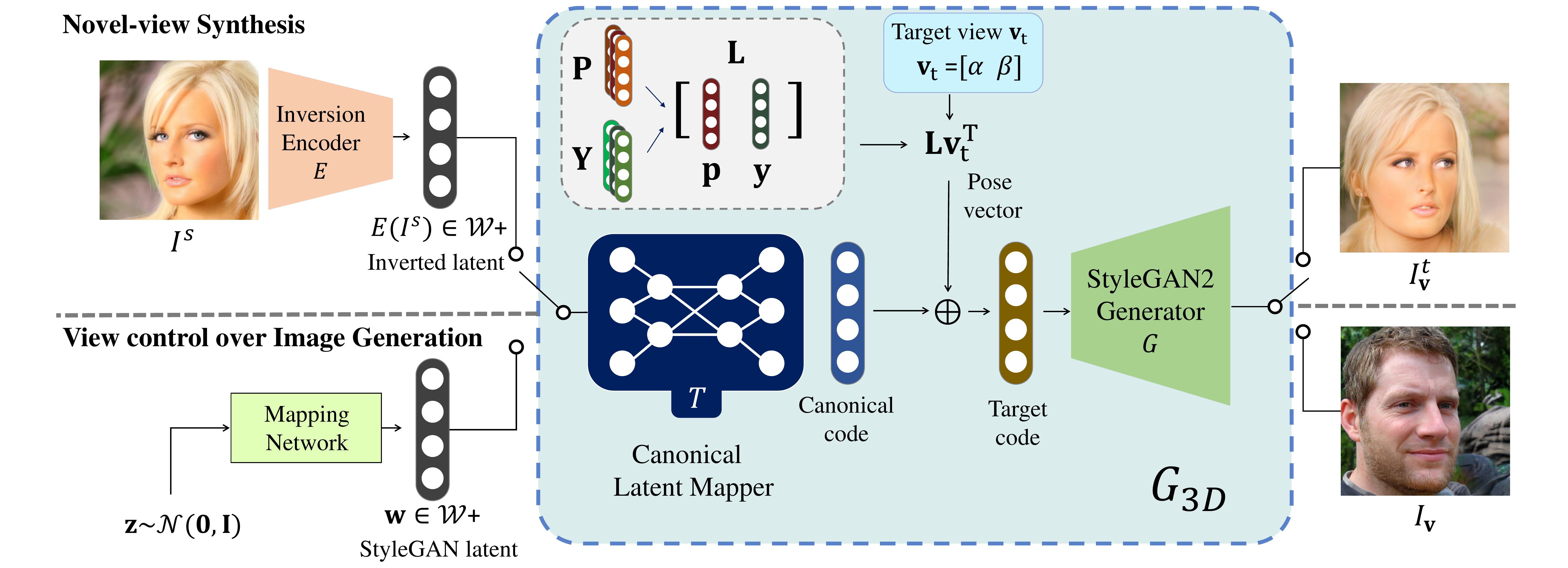}
\caption{\small{The controllable StyleGAN allows explicit control over camera view. It can be used for novel pose synthesis (upper) and view-conditioned image generation (lower).}}
\label{figure:3Dstylegan}
\end{figure} 
Although earlier studies~\cite{shen2020interpreting,nitzan2021large} have shown successful interpolation results with the linear manipulation, unfortunately, they have found sub-optimal solutions that just control the intended pose attribute implicitly rather than explicit control over 3D camera.  
The interesting fact we observed is that the pose-related attribute (e.g., yaw) is not uniquely determined by a single direction. Rather, several orthogonal directions can have different effects on the same attribute. For example, two orthogonal direction A and B both can affect yaw but work differently. 
Based on this observation, we exploit several sub-direction vectors to compensate marginal portion that is not captured by a single direction vector. Our hypothesis is that the optimal direction that follows real geometry can be obtained by a proper combination of the sub-direction vectors. Borrowing the idea of basis in Sec.~\ref{sec:sec3.1}, we construct each of $N$ learnable basis to obtain final pose vectors for pitch and yaw respectively. Therefore, we optimize the matrices $\mathbf{P}=\left[\mathbf{d}^{p}_{1}, \ldots, \mathbf{d}^{p}_{N}\right]$ and $\mathbf{Y}=\left[\mathbf{d}^{y}_{1}, \ldots, \mathbf{d}^{y}_{N}\right]$.  
The process to obtain the target vector can be described as,
\begin{equation} \label{eq:eq10}
    \hat{\mathbf{w}}_{t}
    = \hat{\mathbf{w}}_c + \sum_{i=1}^{N}(\alpha\cdot l^{p}_{i}\mathbf{d}^{p}_{i} + \beta \cdot l^{y}_{i}\mathbf{d}^{y}_{i}),  
\end{equation}
where the $l_i^{p}$ and $l_i^{y}$ represent the learnable scalining factor deciding the importance of basis $\mathbf{d}^{p}_{i}$ and $\mathbf{d}^{y}_{i}$ respectively. To penalize finding redundant directions, we add orthogonal regularization, i.e., 
\begin{equation}
     \mathcal{L}_{\text{reg}} = \left\|\mathbf{P}^{\mathrm{T}} \mathbf{P}-\mathbf{I}\right\|_{1} +\left\|\mathbf{Y}^{\mathrm{T}} \mathbf{Y}-\mathbf{I}\right\|_{1}.
\end{equation}
Similar to the canonical view generation, the model is penalized by the difference of the latent codes ($\mathbf{w}_{t}$ vs. $\hat{\mathbf{w}}_{t}$) and that of the corresponding decoded images ($I'_{t}$ vs. $\hat{I}_{t}$). In addition, we also utilize LPIPS loss. Therefore, the objective function of target view generation is described as,
\begin{equation} \label{eq:eq12}
    \mathcal{L}^{\text{t}} = 
    \lambda_{4}\mathcal{L}^{\text{t}}_{\text{w}} + \lambda_{5}\mathcal{L}^{\text{t}}_{\text{I}} + \lambda_{6}\mathcal{L}^{\text{t}}_{\text{LPIPS}} 
    + \lambda_{7}\mathcal{L}_\text{reg}.
\end{equation}
Finally, the full objective to train the proposed modules can be formulated as $\mathcal{L} = \mathcal{L}^{\text{c}} + \mathcal{L}^{\text{t}}.$
After training, StyleGAN ($G$) becomes a 3D-controllable generator ($G_{3D}$) with the proposed modules as illustrated in Fig.~\ref{figure:3Dstylegan}. We can achieve a high quality image with intended pose by conditioning view as follow,
\begin{equation}
    I_{\mathbf{v}}=G_{3D}(\mathbf{w},\mathbf{v}_t) = G(\mathbf{w}+T(\mathbf{w})+\mathbf{L}\mathbf{v}_{t}^T),
\end{equation}
where $I_{\mathbf{v}}$ represents a generated image with target pose $\mathbf{v}_{t}$ and $\mathbf{w}\in \mathcal{W}+$ is duplicated version of 512-dimensional style vector in $\mathcal{W}$  which is obtained by the mapping network in StyleGAN.
Moreover, we can extend our method to synthesize novel view of real images by combining with GAN inversion, i.e.,
\begin{equation}
    I^{t}_{\mathbf{v}} = G_{3D}(E(I^{s}),\mathbf{v}_{t}),
\end{equation}
where $I^{s}$ is an input source image in arbitrary view and $I^{t}_\mathbf{v}$ denotes a generated target image with target pose $\mathbf{v}_{t}$. Note that our method can handle arbitrary images without exploiting off-the-shelf 3D models such as pose detectors or 3D fitting models. In addition, it synthesizes output at once without an iterative optimization process for overfitting latent code into an input portrait image.      

\subsection{Finding semantic direction with SURF-GAN} \label{sec:sec3.3}
Beyond 3D perception, we can discover semantic directions in the latent space of StyleGAN that can control facial attributes using SURF-GAN generated images. 
Such directions can be obtained by a simple vector arithmetic~\cite{radford2015unsupervised} with two latent codes or  several interpolated samples generated by SURF-GAN.  
Although our approach does not overwhelm state-of-the-art methods analyzing via supervision, it would be a simple yet effective alternative that can provide pose-robust editing directions. Of course, the discovery using SURF-GAN is one of many applicable approaches and we can also utilize the existing semantic analysis methods~\cite{shen2020interpreting,harkonen2020ganspace,shen2021closedform} because 
our model is flexibly compatible with well-studied StyleGAN-based techniques.

\section{Experimental result} \label{sec:sec4}

This section presents qualitative and quantitative comparisons with state-of-the-art methods and analysis of our method.   
Additional experiments and discussions not included in this paper can be found in the supplementary material.

\begin{figure}[!t]
\centering \includegraphics[width=\linewidth]{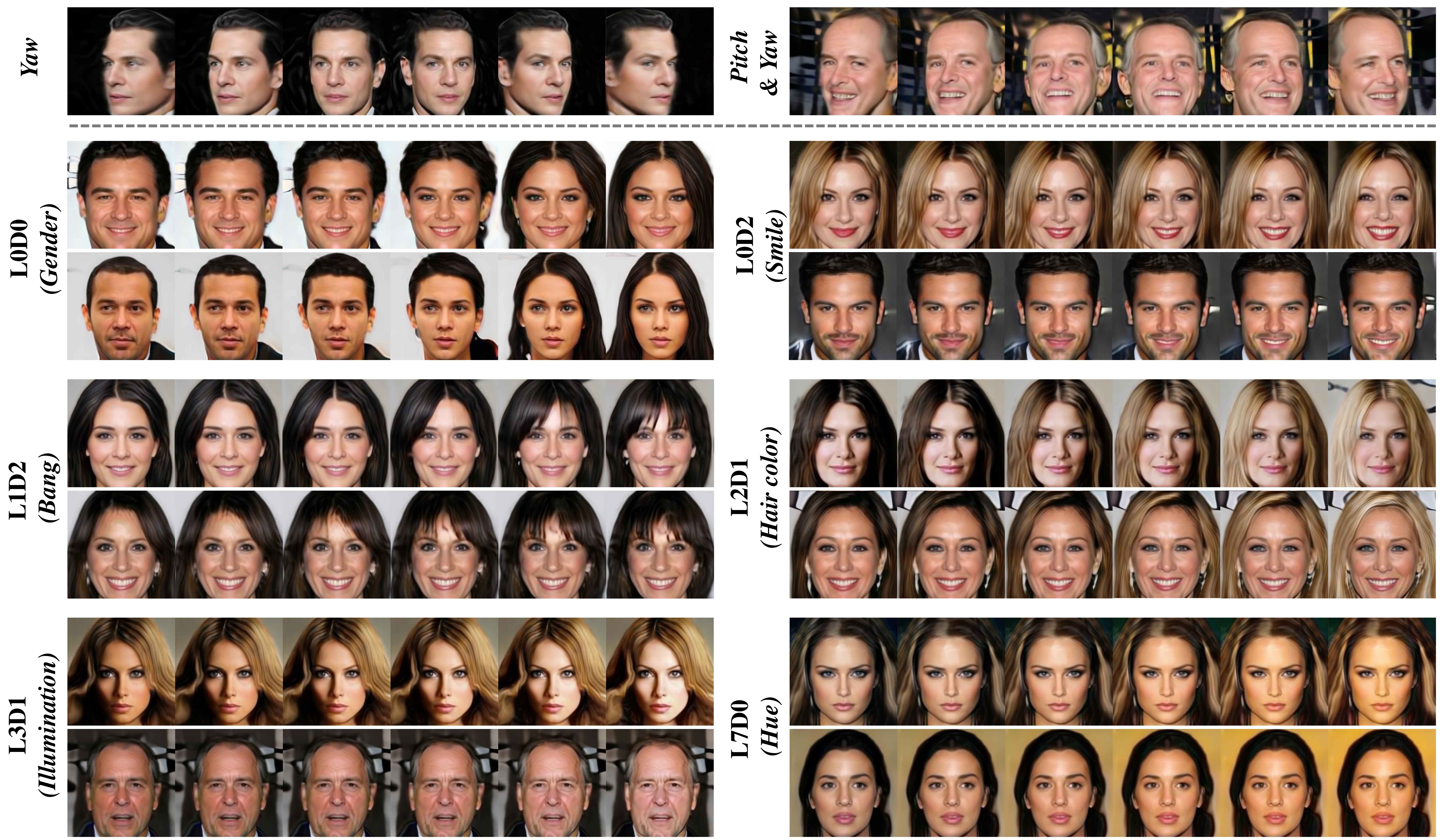}
\caption{\small{Discovered semantic attributes at different layers in SURF-GAN. We can manipulate the attributes (e.g., hair color, gender, etc.) with the control parameters as well as explicit control over 3D camera. L$i$D$j$ denotes the $j^{th}$ basis of the layer $i^{th}$.}}
\label{figure:SURFGAN_semantic} 
\end{figure} 

\subsection{Implementation}
\subsubsection{SURF-GAN.}
We use each of two datasets to train SURF-GAN, i.e., CelebA~\cite{liu2015deep} dataset and FFHQ~\cite{karras2019style} dataset. 
We set the number of sub-modulations in each layer $K$= 6 (Eq.~\ref{eq:eq2} and Eq.~\ref{eq:eq3}) and the number of modulated layers (SURF blocks) is nine ($\therefore t=8$).  
The other settings are roughly the same with those of $\pi$-GAN. More details can be found in the supplementary paper. 

\subsubsection{3D-controllable StyleGAN.}
For training of 3D controllable StyleGAN, we exploit generated images by SURF-GAN trained with FFHQ because StyleGAN~\cite{karras2020analyzing} and GAN inversion encoder~\cite{richardson2021encoding} are pre-trained with FFHQ. 
We design the model to alter only the first four $\mathbf{w}$ vectors (i.e., 4$\times$512) which have been known to control pose~\cite{karras2019style,zhang2020image}.  We set the number of sub-direction $N=5$ (Eq.~\ref{eq:eq10}). The hyper-parameter of the loss function (Eq.~\ref{eq:eq9} and Eq.~\ref{eq:eq12}) are set to $\lambda_{1}$,$\lambda_{4}$=10, $\lambda_{7}=100$, and 1.0 for the others.

\subsection{Controllability of SURF-GAN}

First, we present the attributes of CelebA discovered  by SURF-GAN in Fig.~\ref{figure:SURFGAN_semantic}. As like other 3D-aware GANs~\cite{chan2021pi,Niemeyer2020GIRAFFE,Schwarz2020NEURIPS,zhou2021CIPS3D,gu2021stylenerf,chan2022efficient}, it can synthesis a view-conditioned image, i.e., yaw and pitch can be controlled explicitly with input view direction (top row). 
In contrast to other 3D NeRF-GANs, SURF-GAN can discover semantic attributes in different layers in an unsupervised manner.
Additionally, the discovered attributes can be manipulated by the corresponding control parameters. 
As shown in Fig.~\ref{figure:SURFGAN_semantic}, different layers of SURF-GAN capture diverse attributes such as gender, hair color, illumination, etc. Interestingly, we observe the early layers capture high-level semantics (e.g., overall shape or gender) and the rear layers focus fine details or texture (e.g., illumination or hue). This property is similar to that seen in 2D GANs even though SURF-GAN consists of MLPs without convolutional layers. 
Additional discovered attributes, those of FFHQ and the comparison with $\pi$-GAN, which is a pure NeRF-GAN as like ours can be found in the supplementary material.

\begin{table}[!t] \centering
\caption{\small{Quantitative comparison of the proposed 3D controllable StyleGAN with other 3D controllable generative models. We use FID, pose accuracy, and frames per second for evaluation. $\dagger$ denotes quoting from the original paper. }}
\label{tb1:generative}
\begin{tabular}{c c  c c c c  c}
\toprule
            & ConfigNet   & $\; \pi$-GAN  &\; CIPS-3D  &\;  LiftedGAN   & \; Ours   \\
\midrule
FID ($\downarrow$) &  33.41$^{\dagger}$  &47.68 &6.97$^{\dagger}$ &   29.81$^{\dagger}$ & \textbf{4.72} \textbf{}\\
Pose err.\footnotesize{($\times$$10^{-2})$} ($\downarrow$)       & 9.56 & \textbf{3.81} & 9.12  &  5.52 & 4.24 \\
Frames/Sec. ($\uparrow$)      &  \textbf{345}&  4&  22 &   56 &  72  \\

\bottomrule

\end{tabular}
\end{table}

\begin{figure}[!t]
\centering \includegraphics[width=\linewidth]{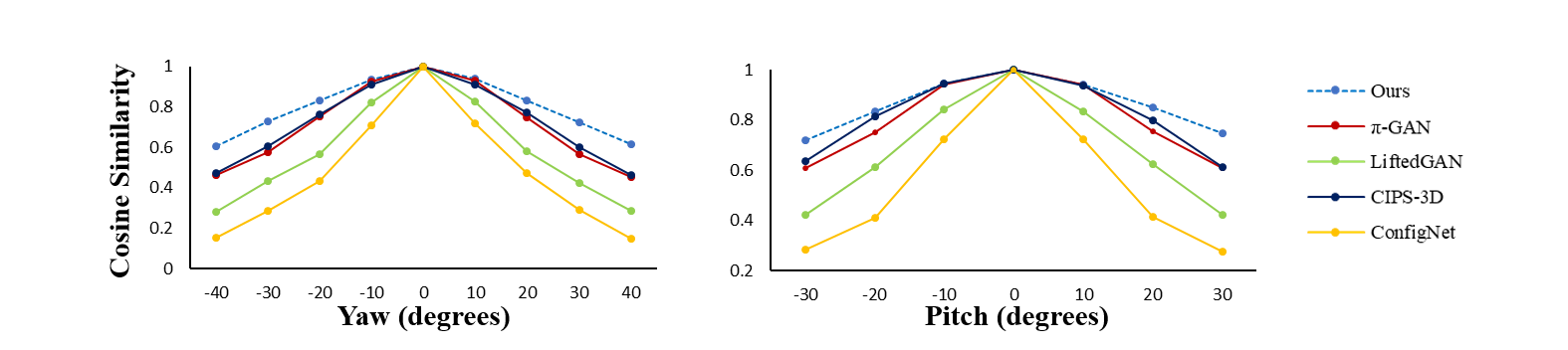}

\caption{Quantitative comparison of 3D-controllable models on identity preservation under different angles using the averaged cosine similarity from ArcFace~\cite{deng2019arcface}.}

\label{figure:grpah_generative} 
\end{figure} 
\subsection{Portrait image generation with 3D control}
\begin{figure}[!t]
\centering \includegraphics[width=\linewidth]{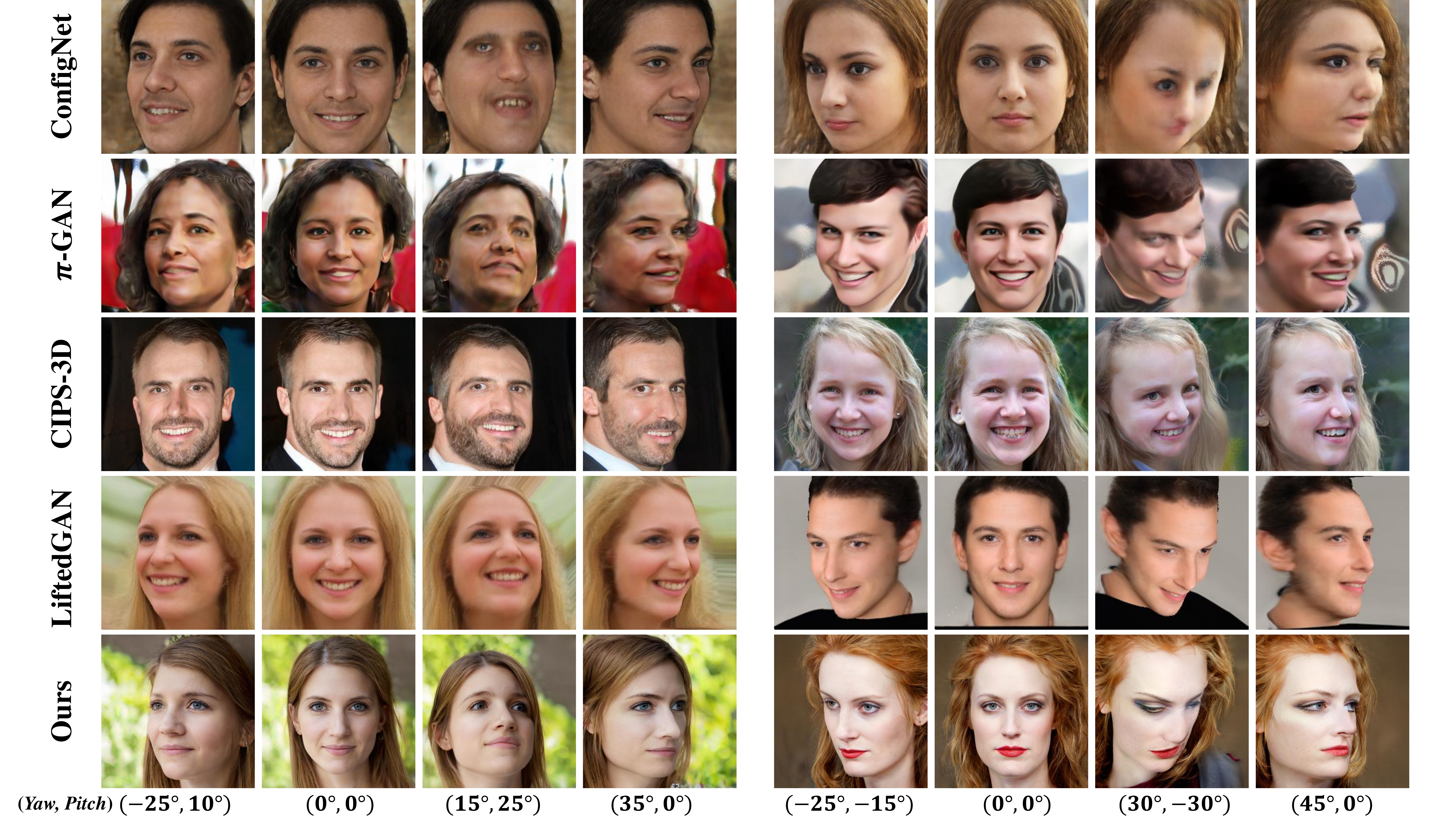}
\caption{\small{Qualitative results of 3D-controllable generative models under target poses.}}
\label{figure:generative}
\end{figure}

To evaluate the performance of the proposed 3D-controllable StyleGAN, we report the qualitative and quantitative comparison with state-of-the-art models~\cite{KowalskiECCV2020,chan2021pi,zhou2021CIPS3D,shi2021lifting} whose generator allows explicit control over pose.
Fig.~\ref{figure:generative} shows synthesis results of each model for given target views. Here, the results are $256^2$ images generated by each method trained with FFHQ~\cite{karras2019style}.
ConfigNet reveals lack of visual quality and weakness in large pose changes. 
$\pi$-GAN shows the accurate geometry because its generator consists of pure NeRF layers, but this property also results in some degenerated visual quality.
CIPS-3D presents improved visual quality by adopting followed 2D INR network, but it suffers from 3D inaccuracy in specific poses.
LiftedGAN generates reasonable outputs according to target views by utilizing differentiable renderer, but it lacks photorealism. Our method generates photorealistic images 
and shows plausible control over pose and multi-view consistency.  
We also report the quantitative comparisons of the models in Table.~\ref{tb1:generative} and Fig.~\ref{figure:grpah_generative}. We use FID score, pose accuracy estimated by 3D model~\cite{zhu2017face}, frames per second, and identity similarity~\cite{deng2019arcface} as evaluation metrics. Compared to 3D-aware models, our method achieves a competitive score on pose accuracy and delivers superior results in efficiency, visual quality, and multi-view consistency. 
Although 2D-based ConfigNet shows overwhelming efficiency, it struggles with multi-view consistency and photorealism.        

\begin{figure}[!t]
\centering \includegraphics[width=12cm]{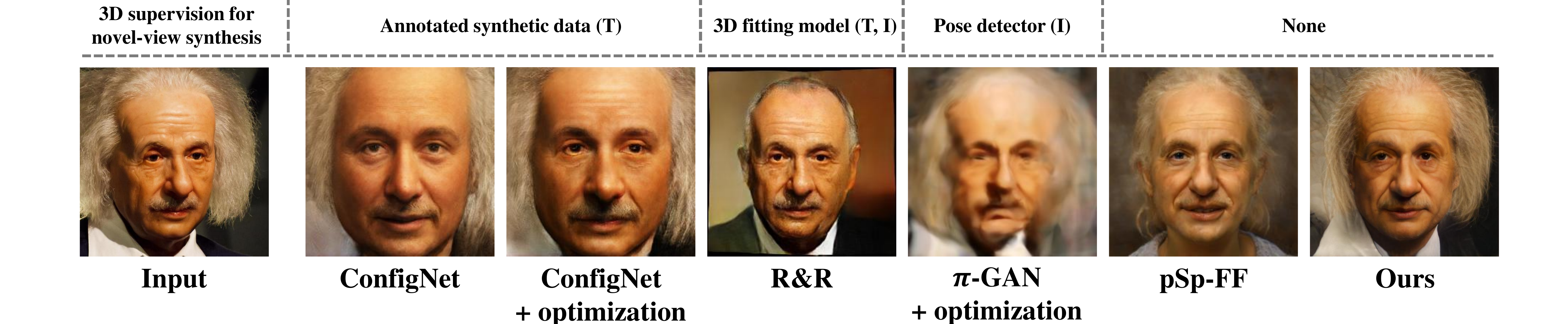}
\caption{\small{Face frontalization results by the methods that can edit the pose of real image. Upper row denotes 3D supervision of each method for training (T) and inference (I).}}
\label{figure:frontal} 
\end{figure} 

\subsection{Novel view synthesis of real image}
\begin{figure}[!t]
\centering \includegraphics[width=\linewidth]{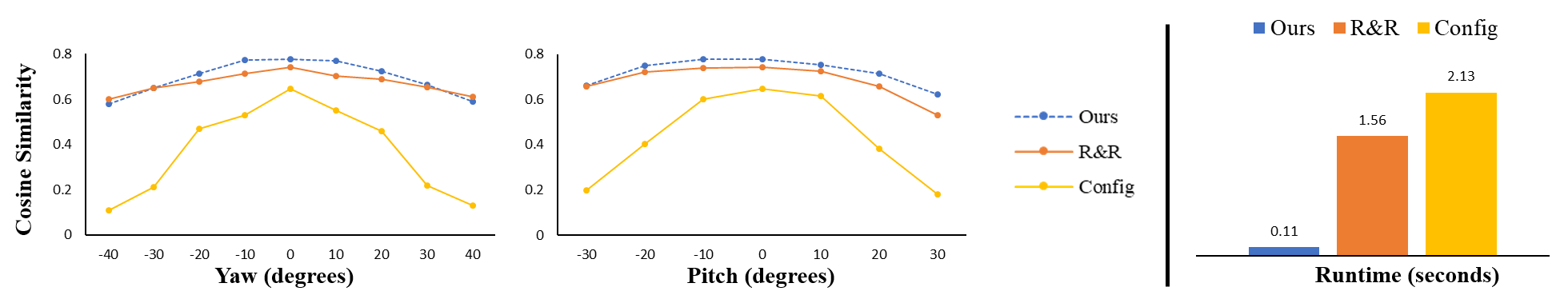}
\caption{\small{Quantitative results of novel view synthesis models. We compute identity similarity between input and synthesized images using ArcFace (left) and runtime (right).  }}
\label{figure:graph_novelview} 
\end{figure}

By utilizing GAN inversion method, our method can perform novel view synthesis from a single portrait. Here, we use pSp~\cite{richardson2021encoding} encoder for the inversion. 
To demonstrate the effectiveness of the canonical mapper, we firstly present the frontalization results in Fig.~\ref{figure:frontal}, which is a special case of novel pose synthesis. 
We mark the 3D supervision in training and inference for each method. Here, pSp-FF denotes the frontalization-only version of pSp~\cite{richardson2021encoding}. Our method successfully generates a canonical view while preserving the identity. 
Next, we further compare the novel view synthesis results of each model. ConfigNet and $\pi$-GAN with optimizing latent code through iterative manner  for overfitting to single test image show inferior results, especially in large pose variation. Rotate-and-Render (R\&R)~\cite{zhou2020rotate} presents reasonable results by exploiting off-the-shelf 3D fitting models~\cite{zhu2017face} in the generation process. However, R\&R loses some fine details of properties of the original, such as hair or background. Our models can edit pose successfully while preserving identity even though it does not require off-the-shelf 3D models and additional optimization for overfitting to an input. It is also demonstrated by the quantitative results in Fig.~\ref{figure:graph_novelview} which reports 
the averaged cosine similarity between input image and outputs at given various angles using ArcFace~\cite{deng2019arcface} and runtime of each method to process a single image.
\begin{figure}[!t]
\centering \includegraphics[width=\linewidth]{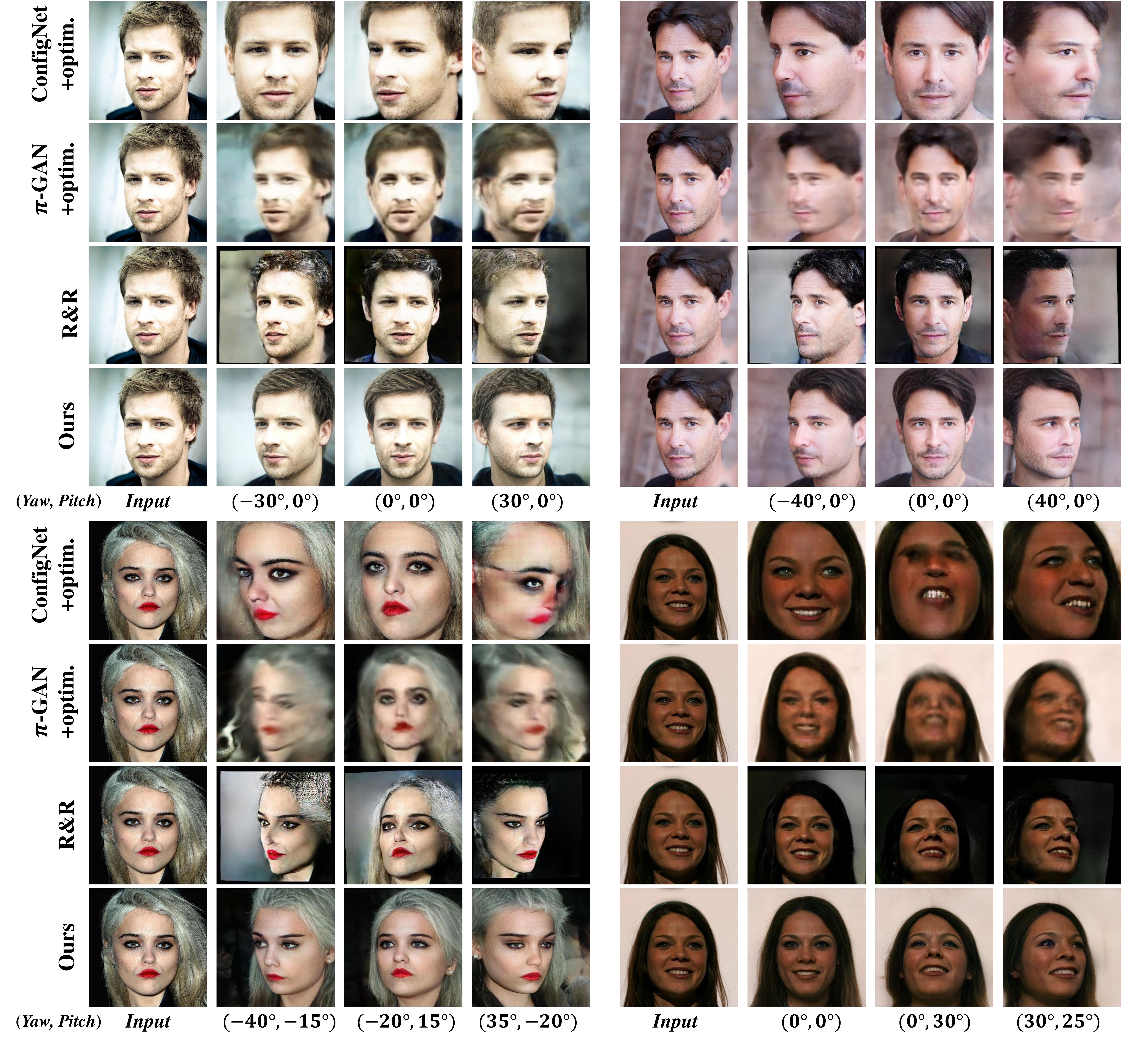}
\caption{\small{Results of novel view synthesis under various target poses using CelebA-HQ.}}
\label{figure:novelview}
\end{figure} 
\subsection{Semantic attribute manipulation under conditioned poses} \label{sec:sec4.5}
\begin{figure}[!t] 
\centering \includegraphics[width=12cm]{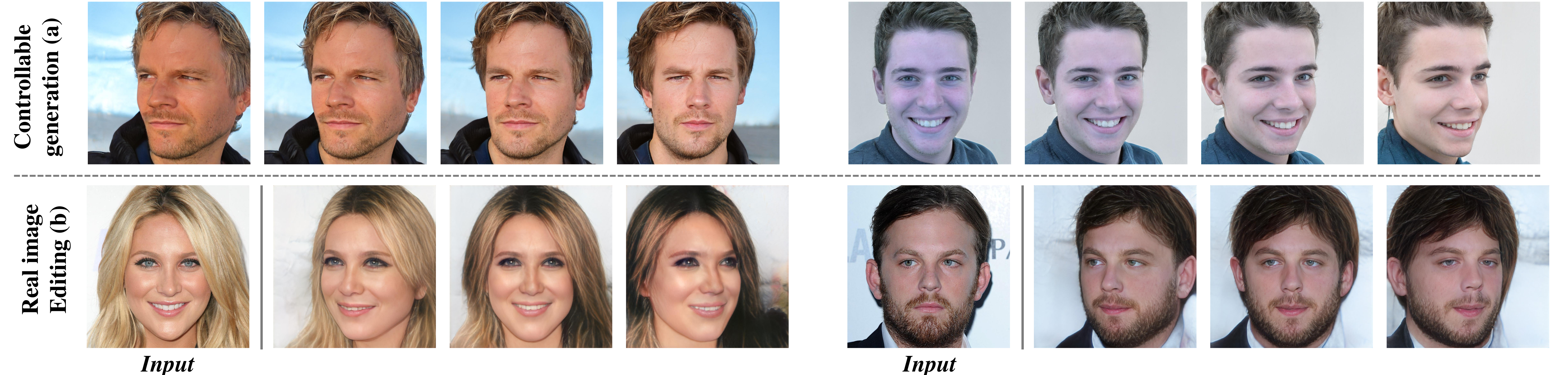}
\caption{\small{Editing both attributes and view direction   by 3D-controllable StyleGAN.}}
\label{figure:edit3D} 
\end{figure}

Fig.~\ref{figure:edit3D} presents the results of semantic attribute editing with pose control  by 3D-controllable StyleGAN. The upper row stands for controllable generation (a) and  the lower represents real image editing (b).   
The presented attributes, i.e., skin color, hue, hair color, and bangs are those discovered by SURF-GAN.

\subsection{Applications}
\begin{figure}[!t]
\centering \includegraphics[width=\linewidth]{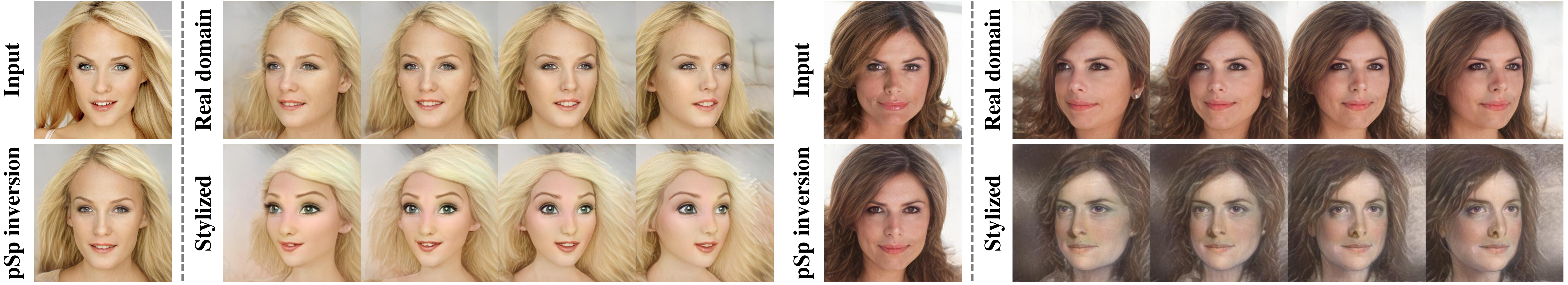}
\caption{\small{Explicit 3D control over real and stylized images.}}
\label{figure:stylization} 
\end{figure}

Our model can be flexibly integrated with other methods that also exploit pre-trained StyleGAN. Beyond the real image domain, we present a novel view synthesis of the stylized images such as toon or painting in Fig.~\ref{figure:stylization}. We use a interpolated StyleGAN proposed by Pinkney and Alder~\cite{pinkney2020resolution} for toonifying  and a transferred StyleGAN trained with MetFace~\cite{Karras2020ada} for painting-style outputs.  

\section{Conclusion}
In this paper, we solved the problems of 3D-aware GANs and 2D GANs by introducing SURF-GAN and 3D-controllable StyleGAN. Unlike other 3D-aware GANs, SURF-GAN can discover meaningful semantics and control them in an unsupervised manner. Using SURF-GAN, we convert StyleGAN to be explicitly 3D-controllable and it delivers outstanding results in both random image generation and novel view synthesis of real image. In addition, our method has the potential to be flexibly combined with  other methods.
We expect our work will be used practically and effectively in various tasks and hope it will open up a new direction in 3D-aware generation and editing fields.
%
\subsubsection{\small{Acknowledgement.}}\small{This work was supported by DMLab. We also thank to Anonymous ECCV Reviewers for their constructive suggestions and discussions on our paper.} 
\clearpage
%
%
\bibliographystyle{splncs04}
\bibliography{egbib}

\clearpage
\title{Injecting 3D Perception of Controllable NeRF-GAN into StyleGAN for Editable \\Portrait Image Synthesis\\ -Supplementary material-}
\author{}
\institute{
}
\titlerunning{Injecting 3D Perception of 3D-aware SURF-GAN into 2D StyleGAN} 
\authorrunning{J. Kwak et al.}

\maketitle

\setcounter{section}{5}
\section{Appendix: SURF-GAN}
In this section, we supplement contents that are not covered in the main paper, i.e., additional details, experiments and discussion about the proposed SURF-GAN.
\subsection{Additional implementation details}
\label{sec:sec6.1}



\subsubsection{Training details.} 
The maximum resolution of SURF-GAN, as well as $\pi$-GAN~\cite{chan2021pi}, which allows stable learning is $64^2$ in our setting, it is because 3D-aware GANs (especially pure NeRF-based GAN) require computationally expensive resources for training. Following $\pi$-GAN, we adopt the progressive growing strategy that the size of the generated image increases progressively. Unlike 2D GANs, the generator does not actually \enquote{grow}. 
Instead, the number of sampled rays increases. Because NeRF-based model can be seen as an implicit continuous function, thus it is theoretically possible to generate arbitrary resolution images. 
Therefore, only the discriminator adds new layers at each stage to handle higher resolutions. We start training at $32^2$ and it is doubled at the next stage. 
In training phase, the control parameter $\mathbf{z}$ is sampled from the standard normal distribution.
The camera pose (pitch and yaw) are sampled from the approximated distribution (normal distribution) of dataset.
We assume a perspective pinhole camera where the field of view (FOV) is $12^{\circ}$. The number of sampled points in each ray is 24 (12 from coarse sampling and 12 from hierarchical sampling). We exploit non-saturating GAN loss with R1 penalty~\cite{mescheder2018training} following $\pi$-GAN. In addition, there is orthogonal regularization of basis ($\mathcal{L}_\text{reg}$) as explained in Sec. 3.1. Finally, pose loss is adopted optionally on different purposes we discuss it in Sec.~\ref{sec:sec6.4}. We adopt ADAM~\cite{kingma2014adam} optimizer with $\beta_1=0$ and $\beta_2=0.99$, and the learning rate is set initially to 0.0001 and it is halved in the next stage.

\subsubsection{SURF-GAN architecture.}
\begin{figure}[!t]
\centering \includegraphics[width=\linewidth]{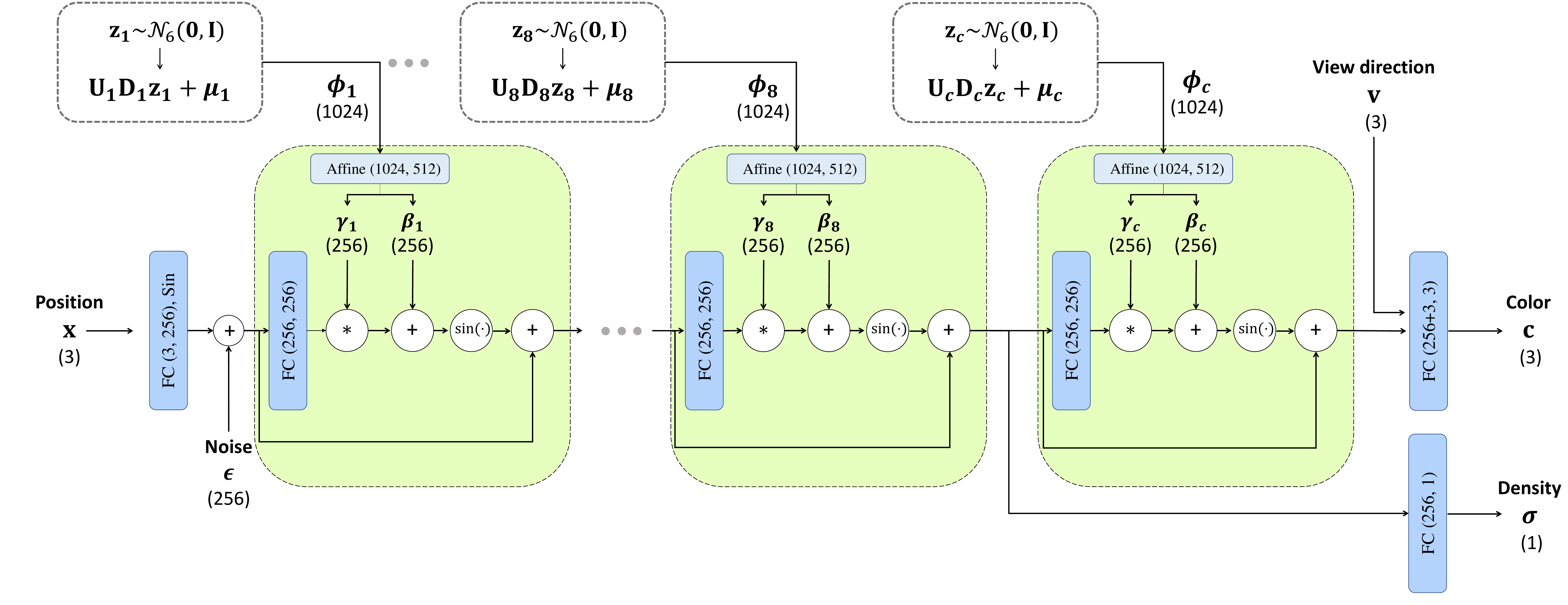}

\caption{Details of SURF-GAN generator.}

\label{figure:fig1_surfgan_generator}
\end{figure}

The architecture of SURF-GAN generator is illustrated in Fig.~\ref{figure:fig1_surfgan_generator}. The discriminator is same with that of $\pi$-GAN~\cite{chan2021pi} except the last layer. Besides the adversarial term that distinguishes real or fake, there is an additional branch that predicts the pose (i.e., pitch and yaw) of an input image. This branch is utilized if the pose loss is adopted, otherwise it is discarded (same as $\pi$-GAN).

\subsection{Comparison with $\pi$-GAN}
We present the comparison results of SURF-GAN with its baseline, $\pi$-GAN (Table.~\ref{tb1:pigan_comparison}). Both approaches belong to pure NeRF-GAN, which consists of NeRF networks without following 2D layers. We evaluate FID score~\cite{heusel2017gans}, pose accuracy, and multi-view consistency of each method. We compute FID score between 50k of generated images and 70k of real images in each dataset. 
Off-the-shelf 3D model (3DDFA~\cite{zhu2017face}) is utilized to evaluate pose accuracy. The reported pose error (Pose err.) is calculated by averaging the difference between target poses and predicted estimated poses. Multi-view consistency (ID) is evaluated  by calculating cosine similarity between canonical view image and others from ArcFace~\cite{deng2019arcface}. Although $\pi$-GAN shows slightly better results in the pose accuracy of CelebA and runtime, SURF-GAN delivers competitive and superior results. For both models, the increased pose error in CelebA is expected to be due to an alignment issue.

\begin{table}[!t] \centering
\caption{Quantitative comparison of SURF-GAN and $\pi$-GAN. Each method is trained on $64^2$ images and the test is conducted on rendered $128^2$ images.}
\label{tb1:pigan_comparison}
\begin{tabular}{ c  c c c c  c}
\toprule
 Method      & \multicolumn{2}{c}{$\pi$-GAN} & \multicolumn{2}{c}{SURF-GAN}      \\ 
 
Dataset                 &\; CelebA  &\; FFHQ  &\; CelebA     & \; FFHQ   \\
\midrule                
FID ($\downarrow$) & $29.54$ & 47.12 &  \textbf{28.88}  &  $\mathbf{44.56}$\\
Pose err.\footnotesize{($\times$$10^{-2})$} ($\downarrow$)        & \textbf{5.81}  & 3.35  & 6.10  & \textbf{2.69}  \\
ID ($\uparrow$) & 0.65 & 0.63 & \textbf{0.68} & \textbf{0.66} \\
Runtime ($\downarrow$)      &  \multicolumn{2}{c}{\textbf{0.10}}    & \multicolumn{2}{c}{0.11}     \\

\bottomrule

\end{tabular}
\end{table}

\subsection{Additional discovered attributes by SURF-GAN}
\begin{figure}[!t]
\centering \includegraphics[width=\linewidth]{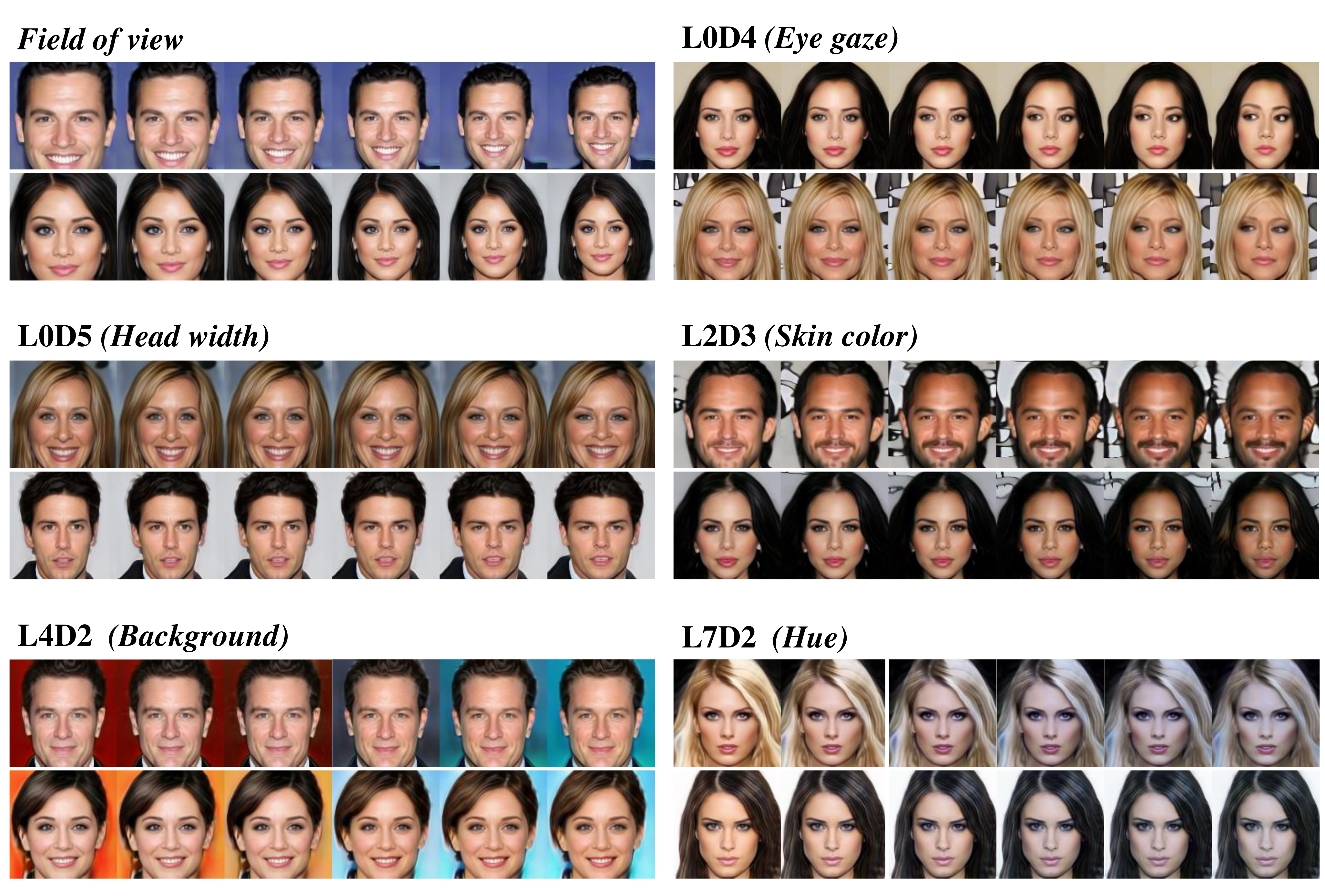}

\caption{Additional attributes discovered by SURF-GAN which are not presented in the main paper (CelebA).}

\label{figure:fig4_surfgan_semantic_celeba} 
\end{figure}

\begin{figure}[!t]
\centering \includegraphics[width=\linewidth]{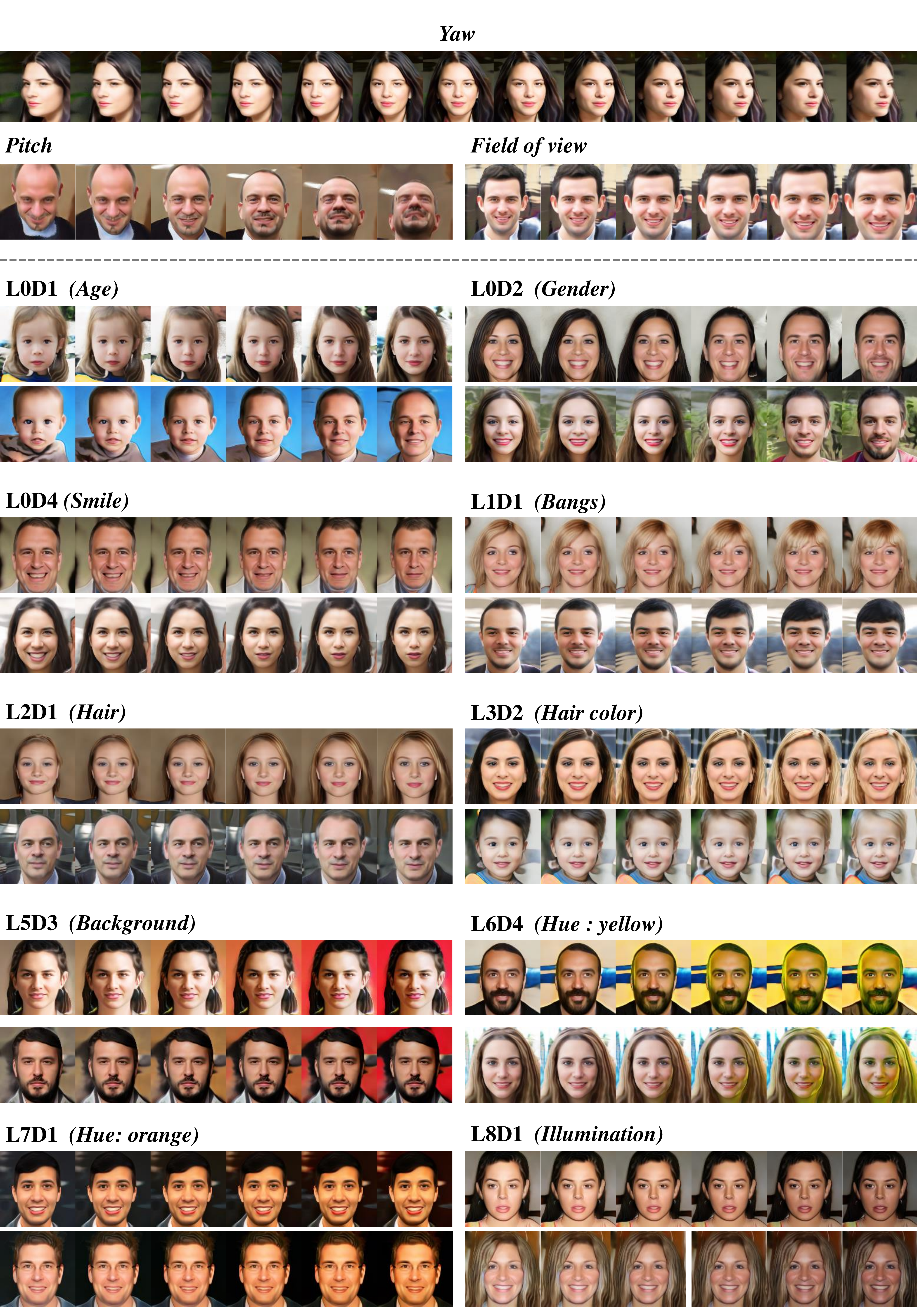}

\caption{Semantic attributes discovered by SURF-GAN when using FFHQ dataset.}

\label{figure:fig5_surfgan_semantic_ffhq} 
\end{figure}

We present the additional attributes in CelebA dataset which are not introduced in the main paper (Fig.~\ref{figure:fig4_surfgan_semantic_celeba}). Note that \enquote{Field of view} is not an discovered attribute, but can be controlled in volume rendering. We also report the semantic attributes of FFHQ in Fig.~\ref{figure:fig5_surfgan_semantic_ffhq}.  

\subsection{Discussion} 
\label{sec:sec6.4}
\subsubsection{Effect of the bottom noise.}
\begin{figure}[!t]
\centering \includegraphics[width=\linewidth]{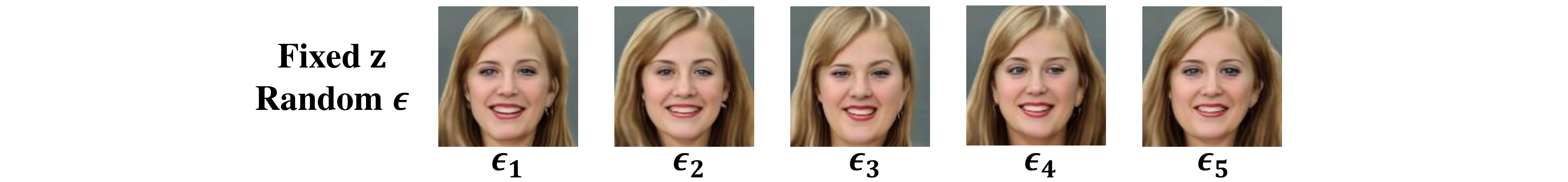}

\caption{The bottom noise $\epsilon$ captures subtle variations ($64\times64$ images).}

\label{figure:supplfig4_bottomnoise} 
\end{figure}

In addition to the layer-wise latent $\mathbf{z}$, our generator also takes the bottom noise $\mathbf{\epsilon}$ as an additional input to capture missing variations (Sec. 3.1). Therefore, the intended role for $\mathbf{\epsilon}$ is to capture the minor variations that have less or not semantic meaning but enhance diversity. Fig.~\ref{figure:supplfig4_bottomnoise} presents generation results when changing only $\mathbf{\epsilon}$. As can be seen, the generator synthesizes the images with minor variation while preserving facial identity.          
\subsubsection{Effect of the progressive growing.}
\begin{figure}[!t]
\centering \includegraphics[width=10cm]{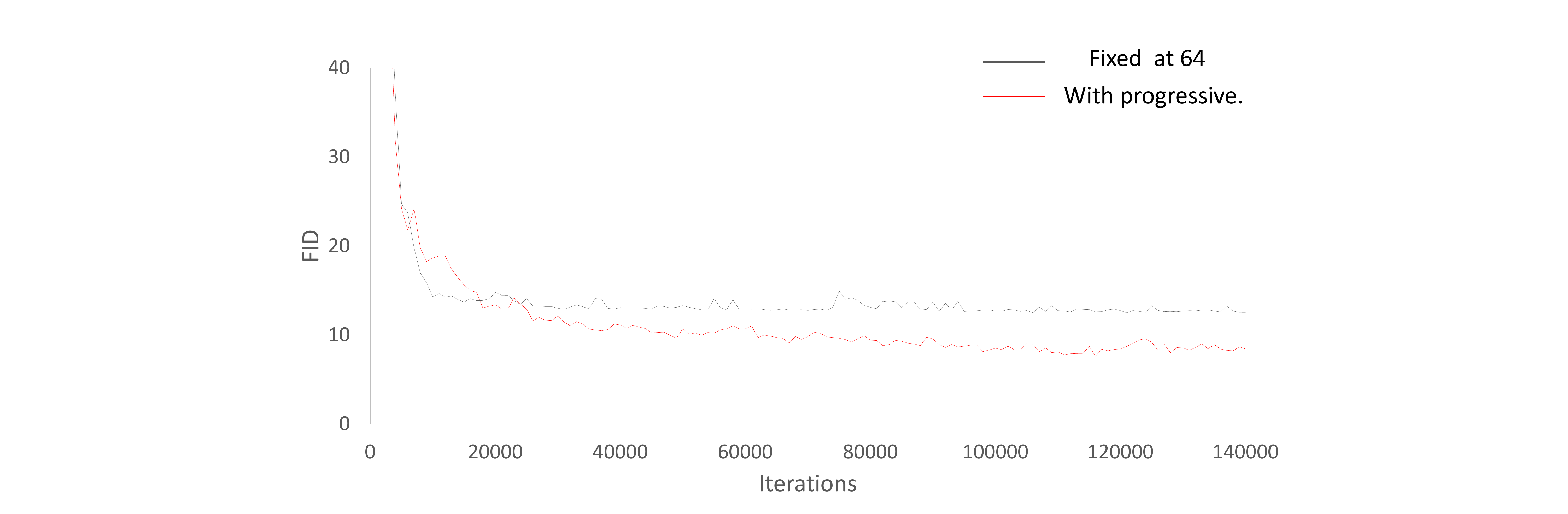}

\caption{FID curve of two variants of SURF-GAN, i.e., with and without progressive growing on CelebA (64$\times$64).}

\label{figure:supplfig5_progressive} 
\end{figure}

As mentioned in Sec.~\ref{sec:sec6.1}, we adopt the progressive growing for training. To demonstrate the effectiveness of the strategy, we report FID score of the variants of our method (i.e., w./ and w.o./ progressive growing) for every 1000 iterations. As can be seen the FID curve in Fig.~\ref{figure:supplfig5_progressive}, there is a gap between two variants. 

\subsubsection{Effect of the pose loss.}
To improve the pose accuracy, we additionally adopt pose loss $\mathcal{L}_\text{pose}$ for training and compare with the original SURF-GAN (w.o./ $\mathcal{L}_\text{pose}$). $\mathcal{L}_\text{pose}$ is calculated as the difference between input viewing directions of generator and those predicted by the discriminator. It is not an adversarial loss, thus both the generator and discriminator learn to minimize the loss. The results are listed in Table.~\ref{tb2:suppltb2_pose_loss}. The introduction of $\mathcal{L}_\text{pose}$ reduces pose error (Pose err.), but sacrifices the visual quality. We exploit this model (w./ $\mathcal{L}_\text{pose}$) for training 3D-controllable StyleGAN (Sec. 3.2) to offer more elaborate pose samples.

\begin{table}[!t] \centering
\caption{Ablation study for training SURF-GAN with and without the pose loss on FFHQ  (128$\times$128).}
\label{tb2:suppltb2_pose_loss}
\begin{tabular}{ c  c c   }
\toprule      &  w.o./ $\mathcal{L}_{\text{pose}}$ &  w./$\mathcal{L}_{\text{pose}}$      \\ 
 \midrule                
FID ($\downarrow$) & \textbf{44.56}   &  45.92\\
Pose err.\footnotesize{($\times$$10^{-2})$} ($\downarrow$)          & 2.69  & \textbf{2.36}  \\

\bottomrule

\end{tabular}
\end{table}

\subsection{Limitation.} 
Although SURF-GAN has several clear advantages such as controllability, there are still inherent limitation as like other 3D-aware GANs. In our model, the color and density of all the points in the rays are calculated independently, thus the amount of computation required to synthesize images increases exponentially as the resolution increases. 
Such issue has been the catalyst for introducing a method of injecting the prior of SURF-GAN into an efficient and expressive StyleGAN2 generator~\cite{karras2020analyzing}. It will be one of our future work to achieve high-resolution with clever and efficient ways, e.g., adopting 2D modules~\cite{chan2022efficient,gu2021stylenerf,or2022stylesdf,deng2022gram}in SURF-GAN generator. 
The other minor limitation is that the same layer does not always capture the same properties for each training. For example, even if a specific dimension (e.g., $L3D2$) of the trained SURF-GAN captures hair color, the same dimension might capture different attributes in the newly trained model. It seems natural because our method is based on unsupervised training, but it makes the process of assigning properties of each layer necessary after training. 

\section{Appendix: 3D-controllable StyleGAN}
This section presents additional experiments and discussion about 3D controllable StyleGAN.  
\subsection{Implementation} 
\label{sec:sec7.1}
\subsubsection{Latent mapper.} The latent mapper consists of five FC layers. It takes an input vector in $\mathcal{W}+$ space and converts it to a canonical vector with the same size.
However, the latent mapper does not edit all elements of $18\times512$ vector, but edits first four style vectors which have known to related to pose~\cite{karras2019style,zhang2020image} (Sec. 4.1), i.e., the input size of the latent mapper is $4\times512$.
Input feature is firstly flatten to 2048-dimensional vector and then converted to intermediate feature $\in \mathbb{R}^{512}$. After going through three intermediate layers, the feature is converted to the canonical vector $\in \mathbb{R}^{4 \times{512}}$ in the last layer.

\subsubsection{Training details.} We leverage SURF-GAN as a multi-view image generator to train 3D-controllable StyleGAN. As described in Sec. 3.2, the poses of source and target images are randomly sampled from pre-defined distribution. In order to train diverse pose angles, we sample pitch and yaw from uniform distributions instead of Gaussian distribution, i.e., the value of pitch and yaw are uniformly sampled from $[-30^{\circ}, 30^{\circ}]$ and $[-45^{\circ}, 45^{\circ}]$, respectively. The resolution of the rendered images is $256^2$ that is same with input size of pSp encoder~\cite{richardson2021encoding}. We use a pretrained StyleGAN2 ($1024^2$) generator for experiments except for MetFace~\cite{karras2018progressive} stylization (here we use a $256^2$ generator for transfer learning).       

\subsection{Discussion}
\subsubsection{Sub-directions.}
\begin{figure}[!t]
\centering \includegraphics[width=\linewidth]{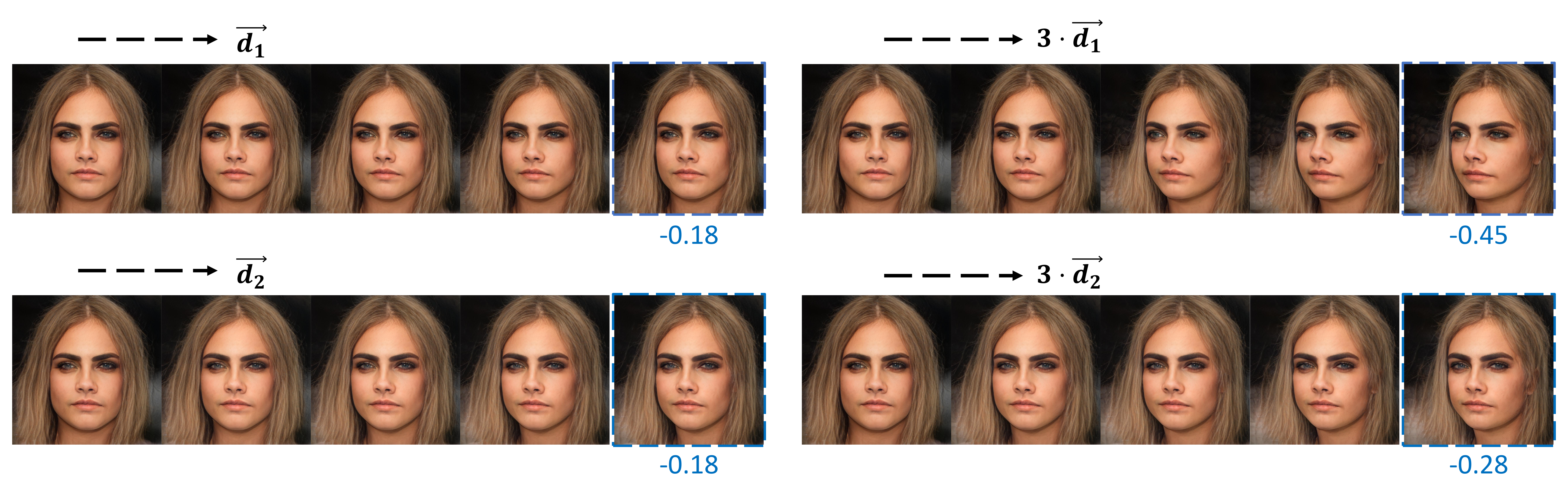}

\caption{Example of non-linearity in pose-related vectors. The blue numbers below the figures indicate the detected poses.}

\label{figure:supplfig_subdirection}
\end{figure} 

To demonstrate the effectiveness of exploiting orthogonal directions (sub-directions) described in Sec. 3.2, we introduce an interpolation example in Fig.~\ref{figure:supplfig_subdirection}. Among learned sub-directions, we select two directions  $\overrightarrow{d_{1}}$ and $\overrightarrow{d_{2}}$ , where both vectors control yaw. As can be seen in the left side of Fig.~\ref{figure:supplfig_subdirection}, they influence almost similarly in small pose variations. However, they shows different interpolation outputs when checking the results by scaling both vectors. It means $\overrightarrow{d_{1}}$ is more involved than $\overrightarrow{d_{2}}$ for generation of images with large pose variations. 

\subsubsection{SURF-GAN-generated images generalization.}
\begin{table}[!t] \centering
\caption{Results of the identity similarity between two decoded images and difference of the estimated poses between SURF-GAN image and the corresponding image decoded by StyleGAN.}
\label{tb3:suppltb3_generalization}
\begin{tabular}{ c  c c   }
\toprule      &   SURF-GAN &  \; Decoded.     \\ 
                
ID  & 0.66   &  0.73\\
\midrule 
Pose diff.     &     \multicolumn{2}{c}{0.003}  \\

\bottomrule

\end{tabular}
\end{table}

We train the latent mapper and the learnable directions using SURF-GAN.
The objective function is calculated with the images decoded by StyleGAN.
However the question that may arise here is \enquote{Can SURF-GAN-generated images be generalized to the training process?}.
To answer the question, we conduct simple experiments. 
First, we measure the cosine similarity of two decoded images at different pose angles using ArcFace~\cite{deng2019arcface} , and also evaluate how much the pose changes in the decoded image using off-the-shelf pose detector.~\cite{zhu2017face}.
The former and the latter are for checking whether identity and pose are maintained, respectively.
Although there is a domain gap between SURF-GAN and StyleGAN, the two images with the same identity in SURF-GAN domain maintain the same identity in StyleGAN domain as can be seen Table.~\ref{tb3:suppltb3_generalization}. 
Moreover, the pose of the SURF-GAN-generated image is hardly changed by the GAN inversion or decoding.  

\subsubsection{Extreme cases.}
\begin{figure}[!t]
\centering \includegraphics[width=\linewidth]{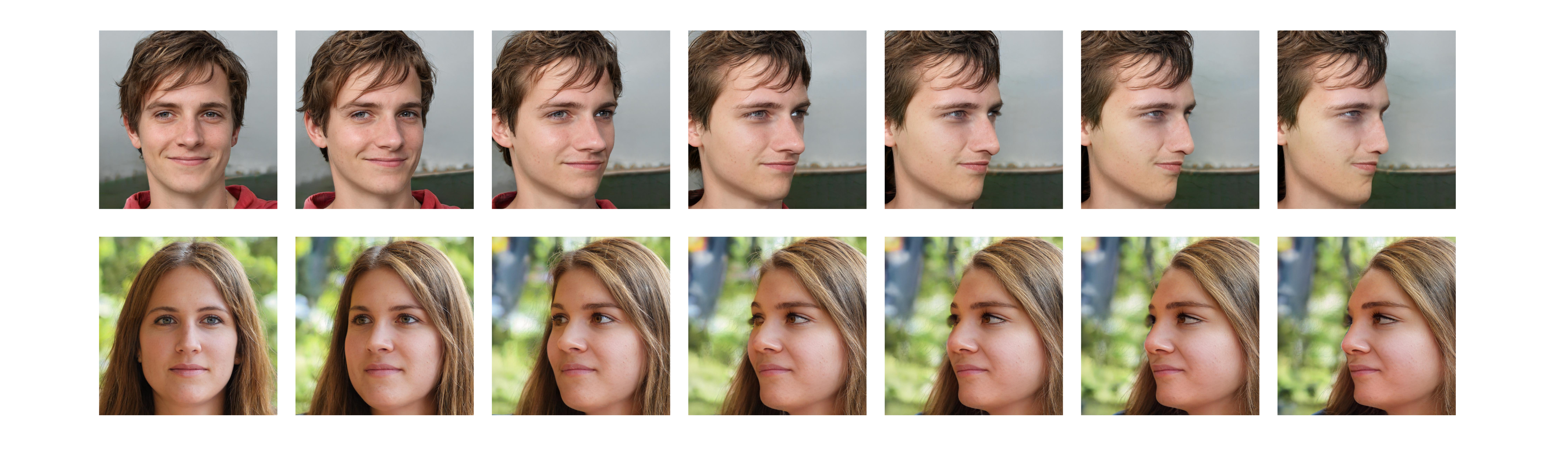}

\caption{Experimental results on extreme poses beyond the range for training.}

\label{figure:supplfig_extreme}
\end{figure} 
As mentioned in Sec.~\ref{sec:sec7.1}, we set the poses for training within a certain range because there are few images with extreme poses in the FFHQ dataset. Nevertheless, we validated the extreme case by giving a large value beyond the pose range as input and observed that there are some cases where plausible images are obtained as shown in Fig.~\ref{figure:supplfig_extreme}.


\subsection{Additional comparison results}
In this subsection, we supplement extra experimental results and discussion to demonstrate the effectiveness of our method. 

\subsubsection{3D-controllable image synthesis.}
\begin{figure}[!t]
\centering \includegraphics[width=\linewidth]{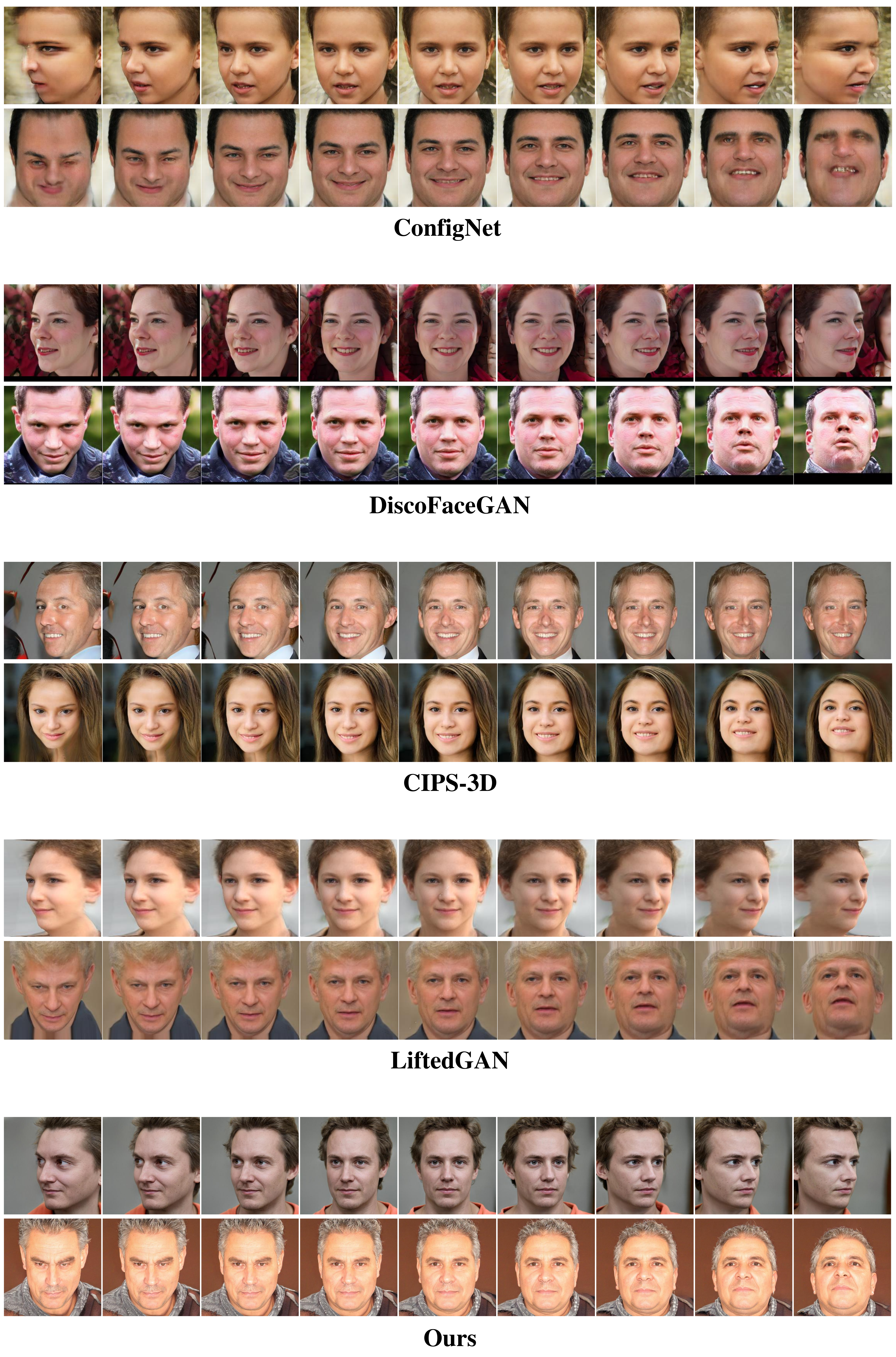}

\caption{Additional qualitative comparison with 3D-controllable generative models.}

\label{figure:grpah_generative} 
\end{figure} 
We first present the additional qualitative comparison with 3D controllable generative models compared in Sec. 4.3. Here we add one more baseline, DiscoFaceGAN~\cite{deng2020disentangled} which is based on 3DMM. Although DiscoFaceGAN does not allow explicit control over the camera pose, it can be implicitly manipulated by an input latent. Therefore, we display the interpolated images by appropriately adjusting the angles of the images at both ends. 

\subsubsection{Novel view synthesis.}
\begin{figure}[!t]
\centering \includegraphics[width=\linewidth]{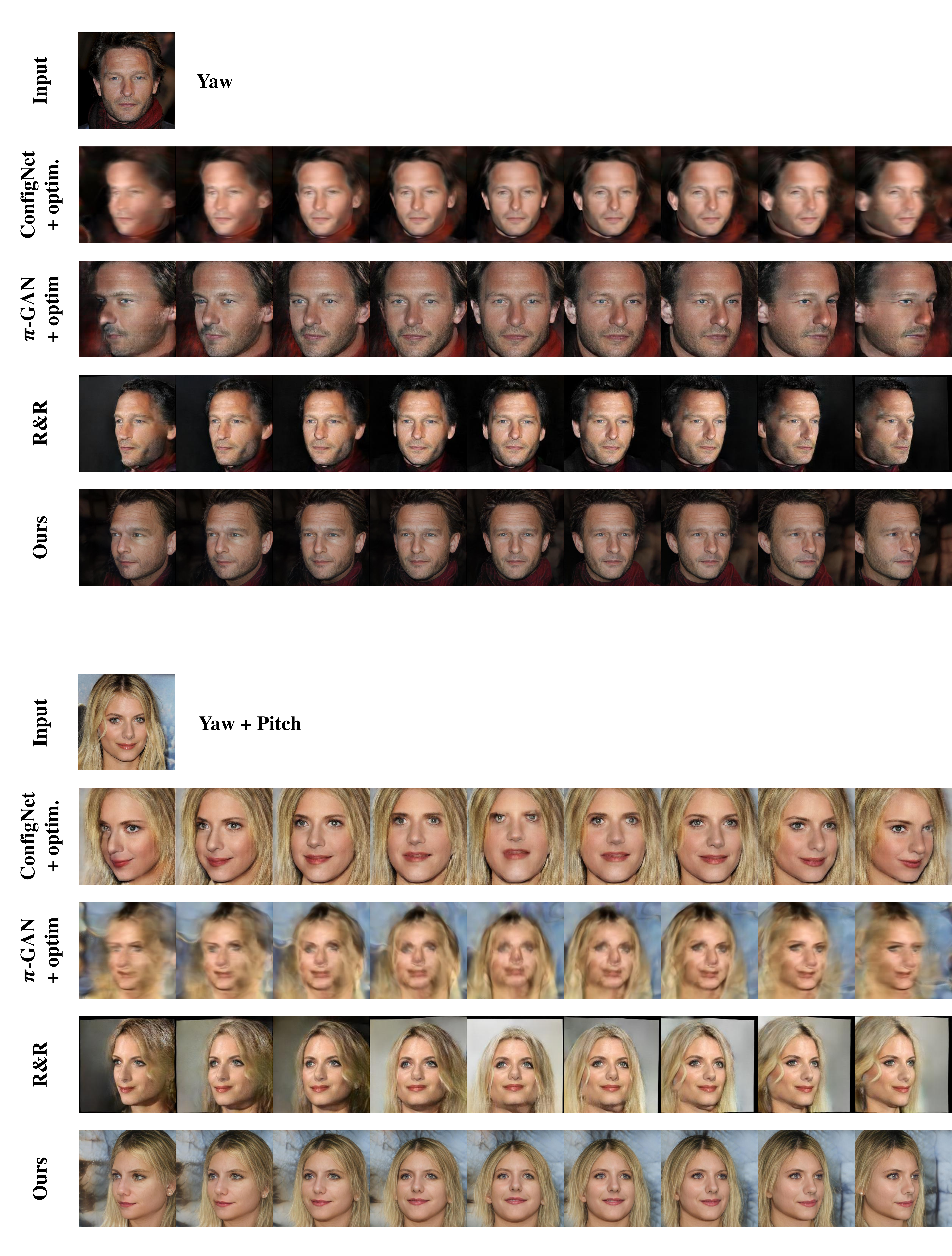}

\caption{Additional qualitative comparison with methods which are capable of novel-view synthesis from a single portrait image.}

\label{figure:supplfig_novelview} 
\end{figure} 

We describe the details not covered in Sec. 4.4 and also present additional qualitative comparison with the competing methods~\cite{chan2021pi,KowalskiECCV2020,zhou2020rotate} for novel view synthesis (Fig.~\ref{figure:supplfig_novelview}). 
For all methods, we use the official implementations provided the authors.

$\pi$-GAN leverages a latent optimization method~\cite{karras2020analyzing} to overfit the latent code to the testing image.  
$\pi$-GAN is a 3D-aware generator and learns 3D geometry from unlabelled 2D images without 3D supervision. However, when it is applied to novel view synthesis, $\pi$-GAN needs camera extrinsics from the testing image to initiate the following iterative optimization (700 iterations). For the camera pose, we exploit off-the-shelf pose detection method~\cite{guo2020towards}. As shown in Fig.~\ref{figure:supplfig_novelview}, the visual quality deteriorates as it deviates from the original pose. It is difficult to generate the radiance field of the target image only with latent code optimization and a small error in the pose estimation greatly affects the result. In addition, it takes a lot of time (164 sec.) to get results for a single image due to the iterative optimization. This is why we excluded $\pi$-gan from the quantitative experiment in Sec. 4.4.  

For ConfigNet~\cite{KowalskiECCV2020}, the real data encoder firstly predicts the latent embedding of a testing image, and then fine-tunes the generator on the image (50 iterations). To handle a real image, it requires pre-processing to align facial images using landmarks from OpenFace~\cite{baltrusaitis2018openface}. Although it shows an overwhelming runtime in random image generation compared to other methods (Sec. 4.3), the runtime drops significantly due to the introduction of the face detection model in novel view synthesis (Sec. 4.4). Note that the reported runtime in Sec. 4.4 (2.13 sec.) does not include the fine-tuning procedure. The whole process takes about 11.25 seconds (9.12 sec. for fine-tuning). Furthermore, ConfigNet struggles to synthesize images with large pose changes. 

Rotate-and-Render (R\&R) is a face rotation method using off-the-shelf 3D fitting network~\cite{zhu2017face} in the overall model, thus it takes some time for 3D fitting (Sec. 4.4). R\&R successfully generates a novel view compared to the previously described methods. However, it loses some details of the original image such as hair or background.

\subsubsection{Comparison with latent-based models.}

\begin{figure}[!t]
\centering \includegraphics[width=\linewidth]{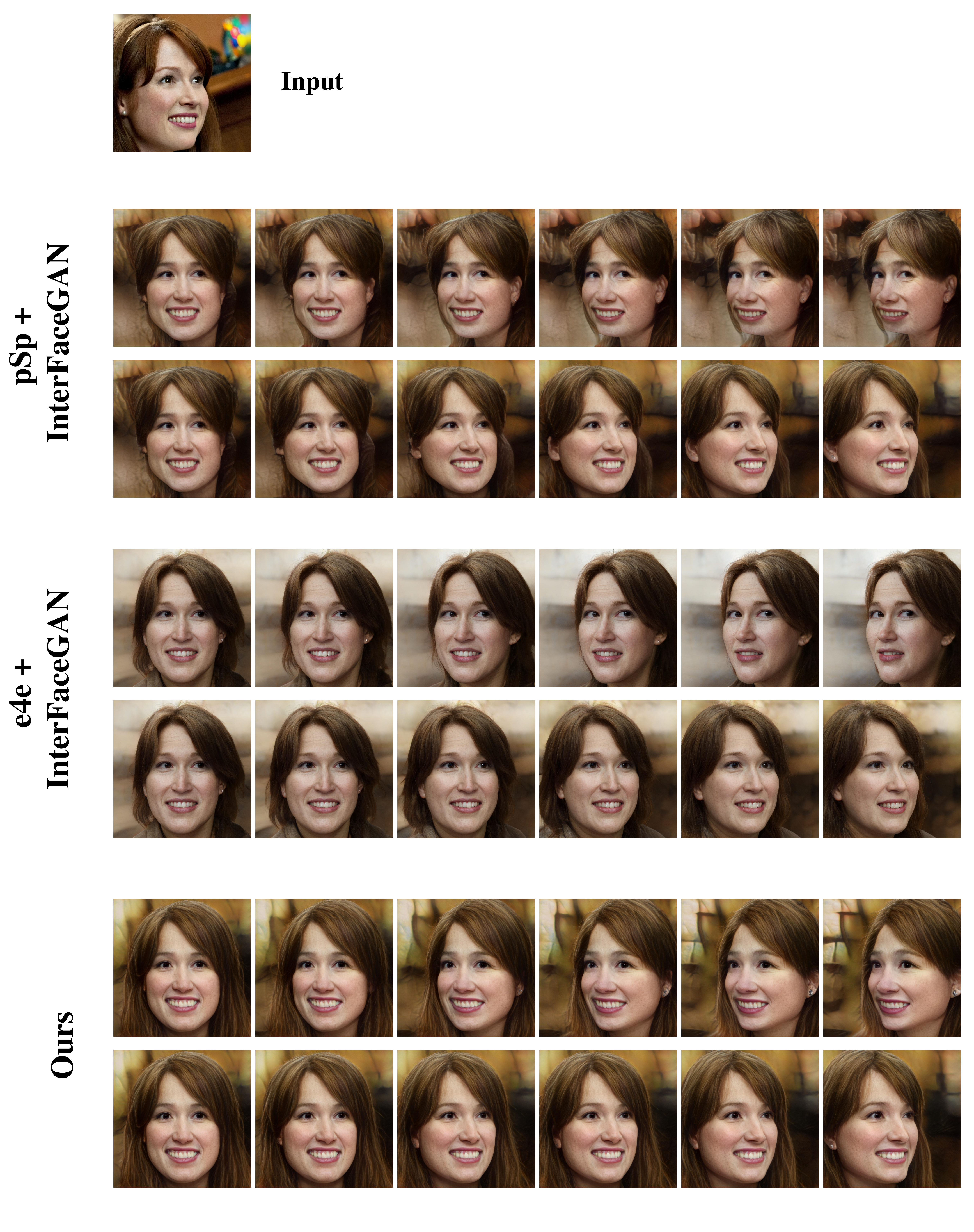}

\caption{Comparison with an existing latent manipulation method (InterFaceGAN).}

\label{figure:latent_compare} 
\end{figure}

The pose editing of our method is based on latent manipulation. We introduce comparison results with the existing latent-based method~\cite{shen2020interpreting} that discovers pose-related direction in the latent space of StyleGAN (Fig.~\ref{figure:latent_compare}). 
Although InterFaceGAN~\cite{shen2020interpreting} successfully disentangles the pose attribute, it requires supervision (landmark) for binary classification of yaw in order to find a semantic hyperplane. 
As Tov et al.~\cite{tov2021designing} have investigated, the results of pose editing with the $\mathcal{W}+$ vectors inverted by pSp~\cite{richardson2021encoding} shows poor editability. It is alleviated by exploiting e4e encoder~\cite{tov2021designing}, but the identity of the input is not well maintained.  
Above all, the important limitation of latent-based models as well as InterFaceGAN is that they only allow implicit control over pose. 
Although it is not unfeasible to generate a target pose using these methods, the process might require a few adjustments to obtain an accurate result.      
Nitzan et al.~\cite{nitzan2021large} have introduced the latent-based linear regression method by showing yaw angle has a linear relationship with the distance from InterFaceGAN's yaw hyperplane. 
Nevertheless, the linearity is not always guaranteed (Fig.~\ref{figure:supplfig_subdirection}), and obtaining the hyperplane requires supervision as mentioned above. There may be an clever alternative to acquire the hyperplane without supervision by leveraging the concept of flipping image~\cite{wu2020unsupervised}, but it can be applied only to yaw, not other properties such as pitch or field of view.   

\clearpage

\subsection{Semantic attribute editing}
\begin{figure}[!t]
\centering \includegraphics[width=\linewidth]{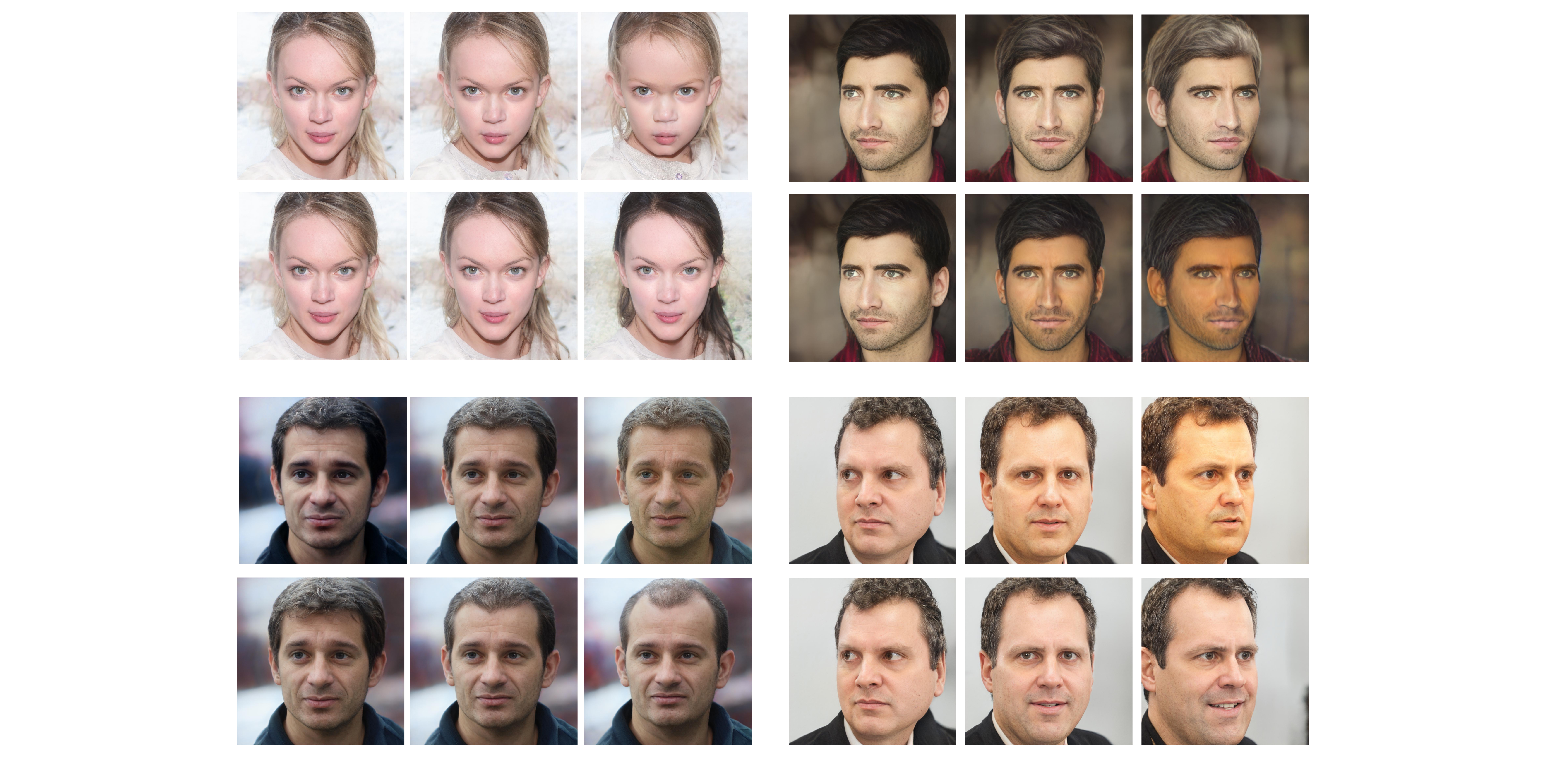}

\caption{Editing results of semantic attributes by calculating the direction vector with images generated by SURF-GAN.}

\label{figure:supplfig_editing} 
\end{figure}

In Sec. 4.5, we presented the results of semantic attribute editing using SURF-GAN, where the direction was calculated by subtracting two inverted SURF-GAN images using pSp encoder~\cite{richardson2021encoding}. However, there is a trade-off between distortion and editability as Tov et al.~\cite{tov2021designing} demonstrated. As a result, some local attributes in Sec. 4.5 are not changed successfully. Although there might be an effect that we use simple vector arithmetic, the main reason is that the $\mathcal{W}+$ space shows weak editability. To address the issue, we investigate the editing results when using e4e~\cite{tov2021designing} encoder to calculate the direction vector and obtain plausible editing results as shown in Fig.~\ref{figure:supplfig_editing}. Note that SURF-GAN samples in Fig.~\ref{figure:fig4_surfgan_semantic_celeba} and Fig.~\ref{figure:fig5_surfgan_semantic_ffhq} are utilized for calculating the directions.      

\begin{figure}[!t]
\centering \includegraphics[width=\linewidth]{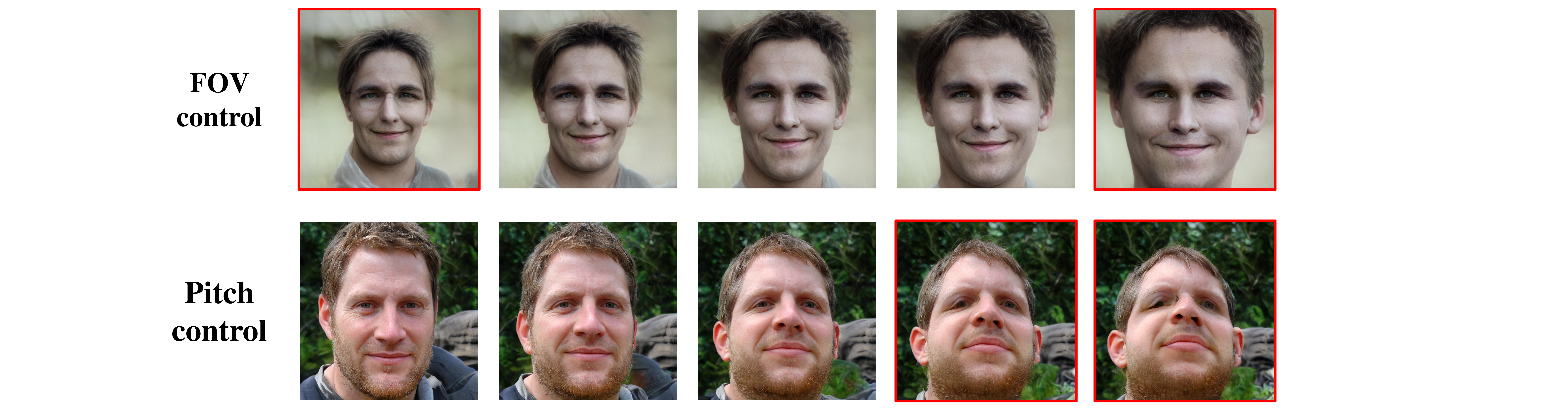}

\caption{Failure cases of our method (red box). For FOV control, we additionally used FOV as a conditional input for training as well as pitch and yaw.}
\label{figure:supplfig_limit} 
\end{figure} 

\subsection{Limitation} 
Although StyleGAN can generate diverse portrait images with high quality, it struggles to generate out-of-distribution images that do not appear in dataset. Therefore, our method also cannot generate those images because our method does not deviate the latent space of StyleGAN. In addition, our method is also affected by the performance of GAN inversion, thus the performance of our model is not guaranteed for images where the inversion method does not work well. We select several failure cases of our method in Fig.~\ref{figure:supplfig_limit}. The other limitation is that as like other pose-disentanlged GANs, our method is not capable of generating 3D representations (e.g., mesh or radiance field). Hence, when it comes to video generation, it shows the problem of \enquote{texture sticking}~\cite{Karras2021} (especially in hair) which is one of the most obvious artifacts in GAN generated videos.
\end{document}